\documentclass[11pt]{article}

\usepackage[preprint]{acl}

\usepackage{times}           
\usepackage[T1]{fontenc}     
\usepackage{inconsolata}     

\usepackage{microtype}
\usepackage{upquote}

\usepackage{amsmath}
\usepackage{amsfonts}

\usepackage{graphicx}

\usepackage{float}
\usepackage{cuted}           

\usepackage[table]{xcolor}   
\usepackage{booktabs}
\usepackage{tabularx}
\usepackage{multirow}
\usepackage{array}
\usepackage{colortbl}
\usepackage{arydshln}
\usepackage{siunitx}

\usepackage{enumitem}
\usepackage{caption}

\usepackage{fontawesome5}
\usepackage{pifont}
\usepackage{contour}
\contourlength{0.1pt}
\newcommand{\cmark}{\textcolor{green!60!black}{\contour{green!60!black}{\ding{51}}}}
\newcommand{\xmark}{\textcolor{red}{\contour{red}{\ding{55}}}}

\usepackage{tcolorbox}
\tcbuselibrary{skins,breakable}

\makeatletter
\let\ALEX@kernelbegin\begin
\let\ALEX@kernelend\end
\makeatother

\usepackage{arabtex} 
\usepackage{utf8}

\makeatletter
\let\begin\ALEX@kernelbegin
\let\end\ALEX@kernelend
\makeatother

\setcode{utf8}
\newcommand{\AR}[1]{\begin{RLtext}#1\end{RLtext}}

\usepackage{epstopdf}

\definecolor{rowgray}{gray}{0.95}
\definecolor{iconcolor}{HTML}{455A64}
\newcommand{\dicon}[1]{\textcolor{iconcolor}{\raisebox{-1pt}{#1}}\hspace{0.6em}}

\definecolor{chatuser}{RGB}{240, 242, 245}
\definecolor{chatagent}{RGB}{235, 248, 255}

\definecolor{egyptprimary}{RGB}{20, 40, 80}
\definecolor{egypthighlight}{RGB}{255, 165, 0}
\definecolor{egyptbg}{RGB}{250, 252, 255}

\definecolor{moroccoprimary}{RGB}{128, 0, 32}
\definecolor{moroccohighlight}{RGB}{255, 215, 0}
\definecolor{moroccobg}{RGB}{255, 250, 250}

\definecolor{mauritaniaprimary}{RGB}{0, 128, 128}
\definecolor{mauritaniahighlight}{RGB}{244, 164, 96}
\definecolor{mauritaniabg}{RGB}{245, 255, 255}

\definecolor{omanprimary}{RGB}{85, 107, 47}
\definecolor{omanhighlight}{RGB}{255, 127, 80}
\definecolor{omanbg}{RGB}{250, 255, 245}

\definecolor{palestineprimary}{RGB}{40, 40, 40}
\definecolor{palestinehighlight}{RGB}{220, 20, 60}
\definecolor{palestinebg}{RGB}{250, 250, 250}

\definecolor{saudiprimary}{RGB}{139, 69, 19}
\definecolor{saudihighlight}{RGB}{210, 180, 140}
\definecolor{saudibg}{RGB}{255, 250, 240}

\newcommand{\activebadge}[3]{%
  \tcbox[on line, arc=2pt, colback=#2!15, colframe=#2,
        fontupper=\bfseries\tiny\color{#1}, boxrule=0.6pt,
        left=1pt, right=1pt, top=1pt, bottom=1pt]{#3}%
}
\newcommand{\passivebadge}[1]{%
  \tcbox[on line, arc=2pt, colback=gray!5, colframe=gray!30,
        fontupper=\bfseries\tiny\color{gray!50!black}, boxrule=0.3pt,
        left=1pt, right=1pt, top=1pt, bottom=1pt]{#1}%
}

\tcbset{
  datacard/.style n args={3}{
    enhanced,
    colback=white,
    colframe=#1,
    fonttitle=\bfseries\small,
    coltitle=white,
    boxrule=0.8pt,
    drop shadow,
    arc=1.5mm,
    left=1.5mm, right=1.5mm, top=1mm, bottom=1mm,
    title={%
      \tcbox[on line, arc=2pt, boxsep=0pt, left=2pt, right=2pt,
            top=1pt, bottom=1pt, colback=white, colframe=white]{\color{#1}\large #3}%
      \hspace{1mm} #2%
    }
  }
}

\newtcolorbox{promptbox}[1][]{
  colback=orange!5,
  colframe=orange!60,
  coltitle=black,
  fonttitle=\bfseries,
  arc=1mm,
  boxrule=0.5pt,
  fontupper=\small\ttfamily,
  left=2mm, right=2mm, top=2mm, bottom=2mm,
  #1
}

\makeatletter
\renewenvironment{abstract}%
  {\centerline{\large\bfseries Abstract}%
   \begin{quote}}%
  {\par\end{quote}}
\makeatother

\usepackage{appendix}

\title{
\raisebox{-1.1ex}{\protect\includegraphics[clip, height=2.5\fontcharht\font`\B]{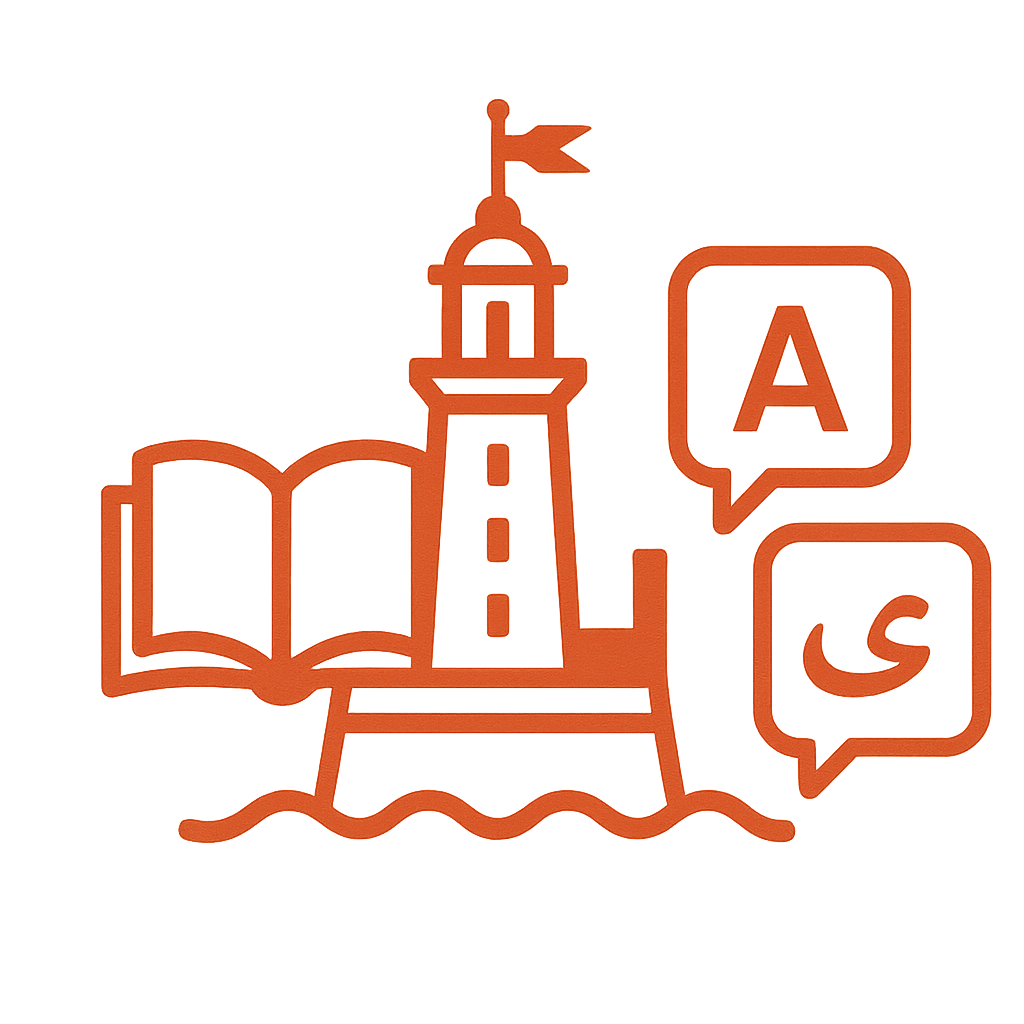}}
Alexandria: A Multi-Domain Dialectal Arabic Machine Translation Dataset for Culturally Inclusive and Linguistically Diverse LLMs
\vspace{0.2em}
}

\author{
  \parbox{\linewidth}{\centering \normalfont
    Abdellah {El Mekki}\textsuperscript{1},
    Samar {M. Magdy}\textsuperscript{1},
    Houdaifa Atou\textsuperscript{3},
    Ruwa AbuHweidi\textsuperscript{4},\\
    Baraah Qawasmeh\textsuperscript{5},
    Omer Nacar\textsuperscript{6},
    Thikra Al-hibiri\textsuperscript{7},
    Razan Saadie\textsuperscript{8},
    Hamzah Alsayadi\textsuperscript{9},\\
    Nadia  {Ghezaiel Hammouda}\textsuperscript{10},
    Alshima Alkhazimi\textsuperscript{11},
    Aya Hamod\textsuperscript{12},
    Al-Yas Al-Ghafri\textsuperscript{11},\\
    Wesam El-Sayed\textsuperscript{13}, 
    Asila {Al sharji}\textsuperscript{11},
    Mohamad Ballout\textsuperscript{8},
    Anas Belfathi\textsuperscript{14},
    Karim  Ghaddar\textsuperscript{8}, \\
    Serry Sibaee\textsuperscript{15},
    Alaa  Aoun\textsuperscript{8},
    Areej  Asiri\textsuperscript{7},
    Lina Abureesh\textsuperscript{4},
    Ahlam Bashiti\textsuperscript{4},
    Majdal Yousef\textsuperscript{4},\\
    Abdulaziz Hafiz\textsuperscript{16},
    Yehdih  Mohamed\textsuperscript{17},
    Emira Hamedtou\textsuperscript{17},
    Brakehe Brahim\textsuperscript{17},\\
    Rahaf Alhamouri\textsuperscript{18},
    Youssef Nafea\textsuperscript{19},
    Aya {El Aatar}\textsuperscript{3},
    Walid Al-Dhabyani\textsuperscript{20, 21},\\
    Emhemed Hamed\textsuperscript{22},
    Sara Shatnawi\textsuperscript{23},
    Fakhraddin Alwajih\textsuperscript{1},
    Khalid Elkhidir\textsuperscript{24},\\
    Ashwag Alasmari\textsuperscript{7},
    Abdurrahman  Gerrio\textsuperscript{22},
    Omar Alshahri\textsuperscript{25},
    AbdelRahim {A. Elmadany}\textsuperscript{1},\\
    Ismail Berrada\textsuperscript{3},
    Amir Azad Adli Alkathiri\textsuperscript{11},
    Fadi {A Zaraket}\textsuperscript{8, 26},\\
    Mustafa Jarrar\textsuperscript{27, 4},
    Yahya Mohamed {El Hadj}\textsuperscript{26, 28},
    Hassan Alhuzali\textsuperscript{16},\\
    Muhammad Abdul-Mageed\textsuperscript{1,2}
  }
  \vspace{0.7em} \\
  \parbox{\linewidth}{\centering \normalfont \small
    \textsuperscript{1}The University of British Columbia,
    \textsuperscript{2}Canada Research Chair in NLP and ML,\\
    \textsuperscript{3}Mohammed VI Polytechnic University,
    \textsuperscript{4}Birzeit University,
    \textsuperscript{5}Western Michigan University,
    \textsuperscript{6}Tuwaiq Academy,\\
    \textsuperscript{7}King Khalid University, 
    \textsuperscript{8}American University of Beirut,
    \textsuperscript{9}Ibb University,
    \textsuperscript{10}University of Hail, \\
    \textsuperscript{11}University of Technology and Applied Sciences,
    \textsuperscript{12}Arab Open University,
    \textsuperscript{13}Minia University,
    \textsuperscript{14}Nantes University,\\
    \textsuperscript{15}Prince Sultan University,
    \textsuperscript{16}Umm Al-Qura University,
    \textsuperscript{17}University of Nouakchott,
    \textsuperscript{18}Fatabyyano,
    \textsuperscript{19}Independent Researcher, \\
    \textsuperscript{20}Hadhramout University,
    \textsuperscript{21}Cairo University,
    \textsuperscript{22}Misurata University,
    \textsuperscript{23}Al-Balqa Applied University,
    \textsuperscript{24}University of Khartoum,
    \textsuperscript{25}Sultan Qaboos Higher Centre for Culture and Science,
    \textsuperscript{26}Arab Center for Research and Policy Studies,\\
    \textsuperscript{27}Hamad Bin Khalifa University,
    \textsuperscript{28}Institut Supérieur du Numérique \\
    \texttt{\{abdellah.elmekki, muhammad.mageed\}@ubc.ca}
  }
}

\begin{document}
\maketitle
\vspace{1cm}

\begin{strip}
  \centering
  \includegraphics[trim={0cm 5.9cm 0cm 0cm}, clip, width=\textwidth]{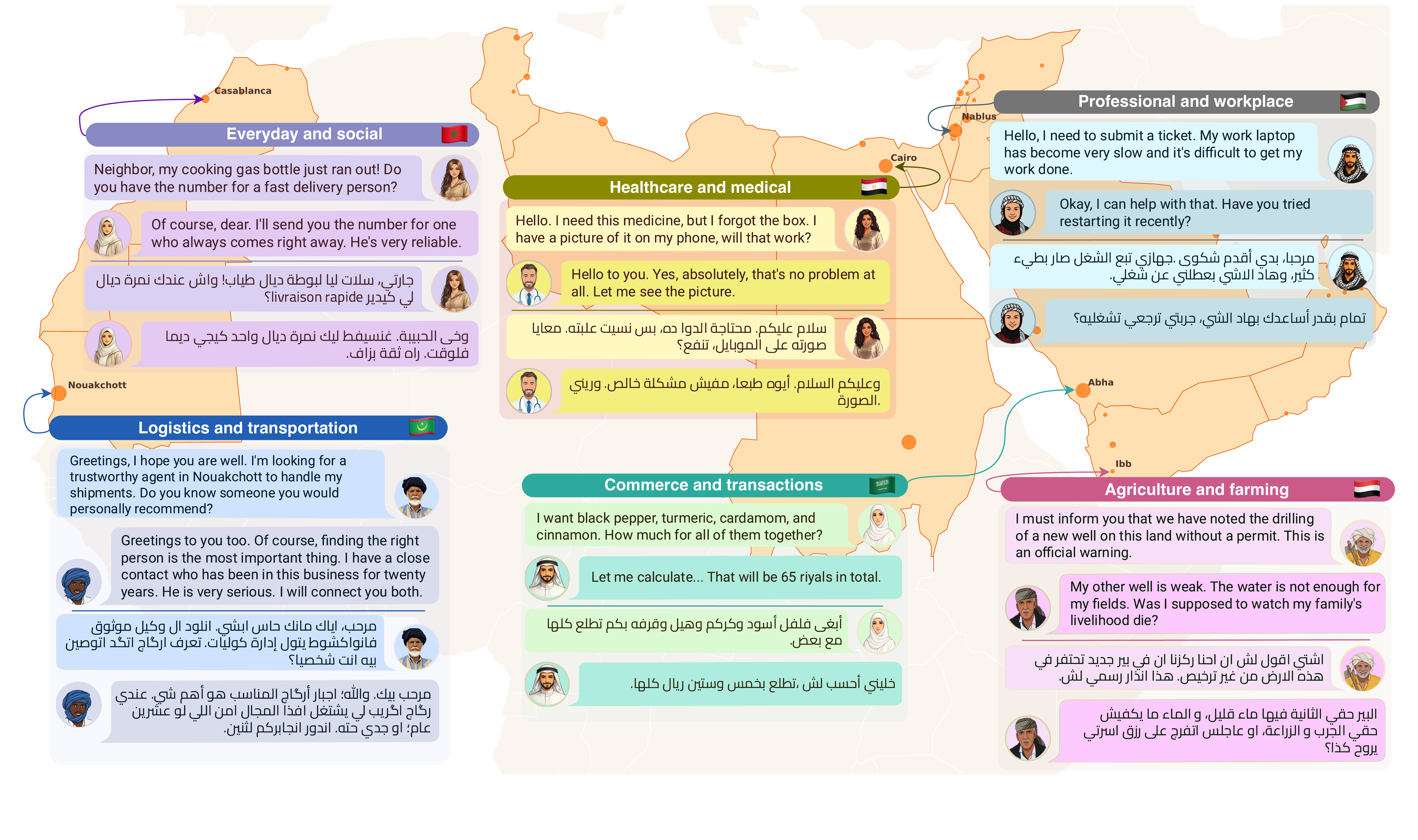}
  \captionof{figure}{Geographic distribution of Alexandria project participants by city across the Arab world. Point diameter is proportional to participant volume. Representative examples (abbreviated to two-turn interactions) are provided to demonstrate the dataset’s coverage across diverse Arabic dialects, domains, and genders.}
  \label{fig:participants_distri}
\end{strip}

\begin{abstract}
Arabic is a highly diglossic language where most daily communication occurs in regional dialects rather than Modern Standard Arabic (MSA). Despite this, machine translation (MT) systems often generalize poorly to dialectal input, limiting their utility for millions of speakers. We introduce \textbf{Alexandria}, a large-scale, community-driven, human-translated dataset designed to bridge this gap. Alexandria covers $13$ Arab countries and $11$ high-impact domains, including \textit{health}, \textit{education}, and \textit{agriculture}. Unlike previous resources, Alexandria provides unprecedented granularity by associating contributions with city-of-origin metadata, capturing authentic local varieties beyond coarse regional labels. The dataset consists of parallel English-Dialectal Arabic multi-turn conversational scenarios annotated with speaker-addressee gender configurations, enabling the study of gender-conditioned variation in dialectal use. Comprising $107$K total turns, Alexandria serves as both a training resource and as a rigorous benchmark for evaluating MT and Large Language Models (LLMs). Our automatic and human evaluation benchmarks the current capabilities of Arabic-aware LLMs in translating across diverse Arabic dialects and sub-dialects while exposing significant persistent challenges.
The Alexandria dataset, the creation prompts, the translation and revision guidelines, and the evaluation code are publicly available in the following repository: \url{https://github.com/UBC-NLP/Alexandria}
\end{abstract}

\section{Introduction}

Machine translation (MT) has evolved from a computational convenience into a critical infrastructure for \textit{digital inclusion}, granting diverse populations access to information, technology, and services. Driven by neural sequence-to-sequence models and large-scale training data, recent advances have substantially improved MT quality for many high-resource language pairs~\cite{tiedemann-thottingal-2020-opus, NIPS2017_3f5ee243, kocmi-etal-2025-command}. Within Arabic, research has also made steady progress on MT involving \textit{Modern Standard Arabic (MSA)}, the lingua franca used in formal writing and broadcast media in the Arab world \cite{Alqudsi2014, nagoudi-etal-2022-turjuman}. Yet, a persistent sociolinguistic gap remains: Arabic is \textit{diglossic} and most everyday communication occurs in regional spoken dialects~\cite{Ferguson01011959, 70ed1286-dfed-3ed2-a29d-fac9677be42d}. These dialects can vary widely across countries and even across cities within the same country \cite{abdul-mageed-etal-2020-toward}, with systematic lexical, morphological, and syntactic differences~\cite{9780199764136.013}. As a result, MT systems trained predominantly on MSA or English-centric resources often generalize poorly to dialectal input, missing vernacular forms and meanings, and thereby limiting the practical utility of MT systems for millions of Arabic speakers~\cite{kadaoui-etal-2023-tarjamat,HARRAT2019262}.

Recent resources aiming to narrow this gap, most notably PADIC~\cite{meftouh-etal-2015-machine} and MADAR~\cite{bouamor-etal-2018-madar}, remain constrained by their design choices and resulting coverage. MADAR is largely organized around travel-and tourism-oriented expressions, while PADIC emphasizes controlled collection and standardized writing practices, improving consistency but limiting naturalistic variation. These choices reduce the extent to which the datasets reflect locally situated usage and fine-grained (e.g., address forms conditioned on gender and social distance, or shifts in register and code choice). 

Similar concerns arise even in widely used multilingual evaluation suites such as FLORES+~\cite{nllbteam2022languageleftbehindscaling}, which has recently served as a standard benchmark for low-resource MT. Prior work reports issues affecting annotation and translation quality in FLORES+~\cite{taguchi-etal-2025-languages}. For Arabic in particular, analyses suggest that some ``dialect'' portions may be overly MSA-leaning (e.g., Moroccan Arabic entries reported as essentially MSA), attenuating the very dialect-specific cues the benchmark aims to test~\cite{abdulmumin-etal-2024-correcting}.

To address this, we introduce \textbf{Alexandria}, a large-scale, human-translated, community-driven dataset designed to capture the richness of \textit{dialectal Arabic} across $11$ domains with \textit{high social impact}, including health, education, agriculture, and finance. Alexandria includes data from \textit{13 Arab countries}, and associates contributions with city-of-origin metadata, moving beyond coarse regional labels (e.g., "Levantine", "North African") to support analyses at finer geographic granularity. 
Alexandria is the outcome of a \textbf{community project} involving \textbf{55 participants} from the \textbf{13 Arab countries}. By involving participants tied to specific cities and local varieties, our collection protocol prioritizes authentic, localized realizations of dialectal forms rather than region-level abstractions. The dataset consists of parallel English-Dialectal Arabic \textbf{multi-turn conversational scenarios} translated to reflect locally relevant contexts. Crucially, each turn is additionally annotated with speaker-addressee gender configuration (e.g., female-to-male), enabling the study of gender-conditioned variation in dialectal language use.

Alexandria, comprising $107$K total turns ($34,488$ conversations), serves two complementary uses for the NLP community: \textbf{(i) Training:} It provides human-translated, domain-diverse conversational data that can be used to train and adapt MT and dialogue-oriented models toward dialectal Arabic in realistic settings. 
\textbf{(ii) Evaluation:} It serves as a benchmark for assessing MT systems and LLM translators under domain, register variation, and speaker-addressee gender configuration conditions, enabling fine-grained analyses of how current models handle dialectal forms and culturally grounded references. To the best of our knowledge, Alexandria is the \textit{first and largest project of its kind}, offering unprecedented granularity in terms of domains, gender annotation, register levels, and city-specific dialectal diversity. By grounding machine translation in the realistic scenarios of the Arab world, we aim to make language technology more accessible, accurate, and culturally inclusive.

\section{Related Work}
\begin{table*}[t]
\centering
\resizebox{\textwidth}{!}{%
\begin{tabular}{lcccccccccccc} 
\toprule
\textbf{Dataset} & \textbf{\# Sentence Pairs / Turns} & \textbf{\# Dialects (countries)} & \textbf{Granularity} & \textbf{Src Type} & \textbf{Direction} & \textbf{\# Domains} & \textbf{Avg. words} & \textbf{Distinct-2} & \textbf{LC} & \textbf{CS} & \textbf{GD} & \textbf{PR} \\
\midrule
PADIC~\cite{meftouh-etal-2015-machine} & 38K & 6 & Country & Sentence & Eng $\leftrightarrow$ Dialect & 1 & 6.77 & 0.782 & \xmark & \xmark & \xmark & \xmark \\
MADAR~\cite{bouamor-etal-2018-madar} & 100K & 13 & City & Sentence & Eng $\leftrightarrow$ Dialect & 1 & 5.73 & 0.768 & \xmark & \xmark & \xmark & \xmark \\
FLORES+~\cite{nllbteam2022languageleftbehindscaling} & 16K & 9 & Country & Sentence & Eng $\leftrightarrow$ Dialect & 3 & 18.39 & 0.898 & \xmark & \xmark & \xmark & \xmark \\
\midrule
\textbf{Alexandria (ours)} & \textbf{107K} & \textbf{13} & \textbf{City} & \textbf{Multi-turn} & \textbf{Eng $\leftrightarrow$ Dialect} & \textbf{11} & \textbf{13.23} & \textbf{0.826} & \textbf{\cmark} & \textbf{\cmark} & \textbf{\cmark} & \textbf{\cmark} \\
\bottomrule
\end{tabular}%
}
\caption{Comparison of Alexandria against existing parallel datasets for Arabic dialects. ($LC=$ Local Context; $CS=$ Code-Switching; $GD= $ gender-direction annotations; $PR=$ persona roles).}
\label{tab:alexandria_comparison}
\end{table*}

\paragraph{Arabic Diglossia and the MT Gap.}  
Despite significant progress in Neural MT, Arabic continues to struggle with the challenges of diglossia and data scarcity~\cite{zbib-etal-2012-machine, sajjad-etal-2020-arabench, kadaoui-etal-2023-tarjamat}. Early parallel corpora remain limited in scale: PADIC \cite{meftouh-etal-2015-machine} provides approximately 6,400 parallel sentences per dialect across five Maghrebi and Levantine varieties, while MADAR \cite{bouamor-etal-2018-madar} translates $2,000$ sentences into $25$ city-specific dialects. Even the FLORES+ benchmark \cite{nllbteam2022languageleftbehindscaling} includes fewer than $1,000$ dev/test sentences per $9$ covered Arabic dialects, leaving many dialects underrepresented. Recent efforts, such as WMT24++ \cite{deutsch-etal-2025-wmt24}, have introduced human-written references and post-edits for several languages, including Egyptian and Saudi Arabic. Furthermore, existing benchmarks are often limited by narrow domains, short sentence lengths, and a lack of context-sensitive translations \cite{malaysha-etal-2024-arafinnlp, abdulmumin-etal-2024-correcting, taguchi-etal-2025-languages}. Furthermore, recent work has addressed gender-aware MT for Arabic~\cite{8374387, alhafni-etal-2022-arabic} to mitigate gender bias; however, these efforts have been limited to MSA. Our Alexandria dataset addresses these gaps by providing a large-scale corpus covering $13$ Arab countries and $11$ domains, featuring conversation-based contexts and granular city-level metadata. Table~\ref{tab:alexandria_comparison} contrasts Alexandria with existing datasets across dialect coverage, domain diversity, and annotation strategies.

\paragraph{Arabic-Capable LLMs and Evaluation.} The rise of LLMs has shifted research focus toward adapting models to specific communities to better reflect their unique linguistic and cultural nuances. This has prompted the release of several evaluation datasets, such as Palm~\cite{alwajih-etal-2025-palm}, Pearl~\cite{alwajih-etal-2025-pearl}, and AraDice~\cite{mousi-etal-2025-aradice}, which assess diverse cultural dimensions and modalities. In terms of modeling, several methodologies have recently emerged to showcase the adaptation of LLMs to the Arab world, notably NileChat~\cite{el-mekki-etal-2025-nilechat} and Fanar~\cite{fanarteam2025fanararabiccentricmultimodalgenerative}. We position Alexandria as a vital contribution to this landscape; it serves not only as a benchmark but also as a powerful tool for the adaptation of conversational LLMs tailored to the specific needs of the Arab world.

\section{Alexandria Dataset Creation}

\begin{figure*}[!ht]
    \centering
    \includegraphics[width=0.99\textwidth]{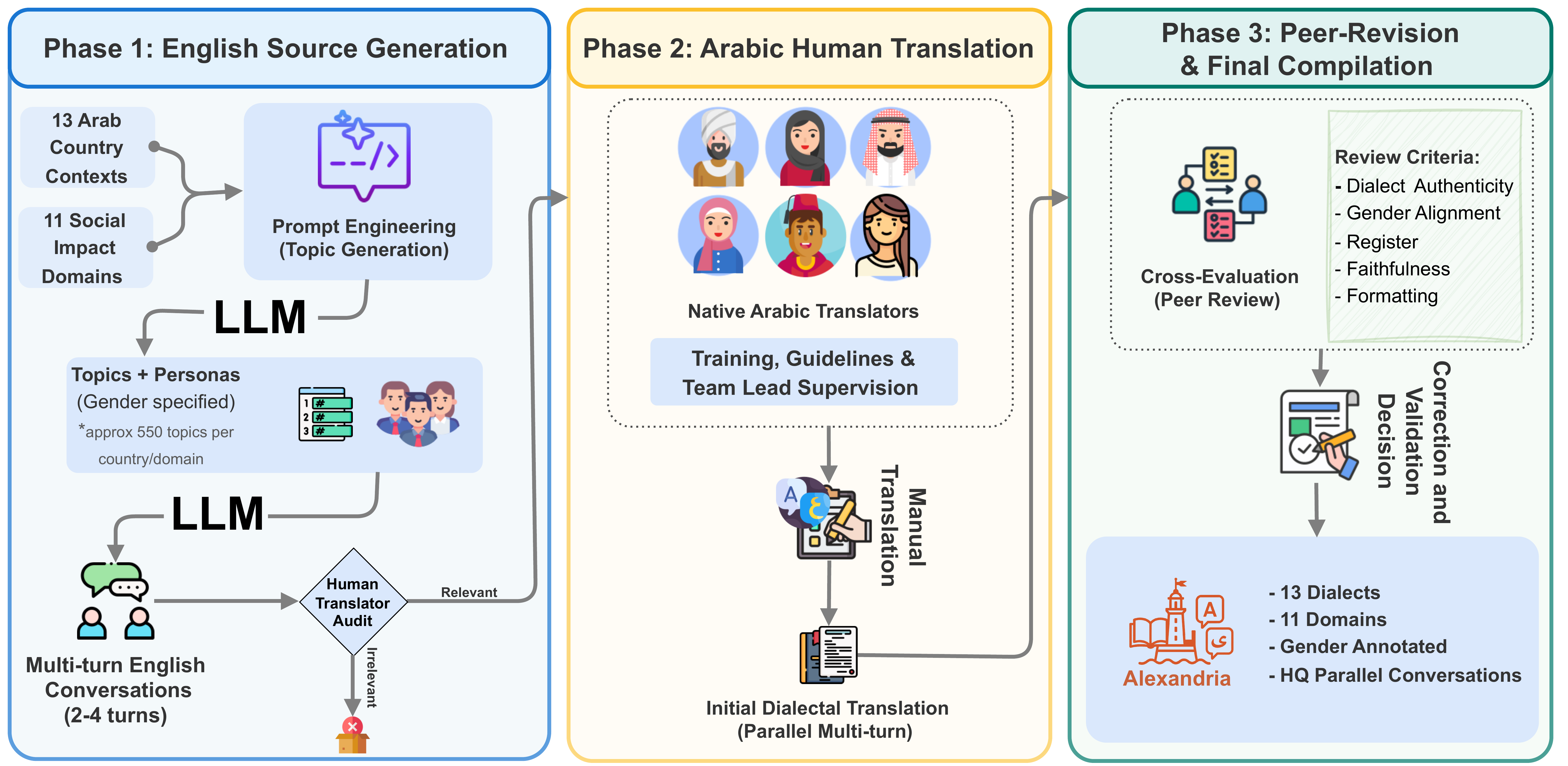}
    \caption{The data creation workflow for the Alexandria dataset. The process illustrates three key phases: \textit{(i)} English source generation, \textit{(ii)} human translation into Dialectal Arabic, and \textit{(iii)} peer-revision and correction.}
    \label{fig:alex_workflow}
\end{figure*}

The Alexandria dataset was created through a six-month, community-driven effort involving 55 team members ($29$ women, $26$ men)\footnote{Self-reported; binary categories} from $13$ Arab countries\footnote{Egypt (EG), Jordan (JO), Lebanon (LB), Libya (LY), Mauritania (MR), Morocco (MA), Oman (OM), Palestine (PS), Saudi Arabia (SA), Sudan (SD), Syria (SY), Tunisia (TN), Yemen (YE).}. Participants were involved to represent local, city-anchored dialectal varieties; the full list of covered sub-dialects appears in Table~\ref{tab:arabic_dialects} (Appendix). Each country team was coordinated by a country lead, who supported member on-boarding and localized annotation guideline examples while preserving a shared annotation schema across dialects. We employed a structured coordination process with weekly checks to ensure consistent progress.

The dataset consists of turn-aligned parallel multi-turn conversations between English and dialectal Arabic, spanning across $11$ domains relevant to public service and everyday life.\footnote{We include domains from the set \textit{\{Agriculture/Farming, Commerce/Transactions, Construction/Real Estate, Education/Academia, Energy/Resources, Everyday/Social, Healthcare/Medical, Legal/Financial, Logistics/Transportation, Professional/Workplace, Tourism/Hospitality\}}.} Additionally, conversations are constructed around persona profiles and include speaker-addressee gender configurations as metadata. Figure \ref{fig:alex_workflow} illustrates our workflow for creating Alexandria. In the following sections, we describe each phase in detail.

\subsection{Alexandria English Sources}
\subsubsection{English Source Generation} 
To address common limitations in prior dialectal Arabic MT resources, such as short utterances, narrow topical coverage, and limited conversational context~\cite{bouamor-etal-2018-madar}, we construct Alexandria using a controlled source-generation pipeline. Specifically, we use \texttt{Gemini-2.5 Pro}~\cite{comanici2025gemini25pushingfrontier} to generate multi-turn English conversational scenarios conditioned on the target country and domain. The process is carried out in two phases to promote topical diversity and minimize near-duplicate conversations. 

\paragraph{Phase 1: Topic Creation.} For each country-domain pair, we use a shared prompt template (with country- and domain-specific variables) to generate 550 topic specifications spanning diverse personas and scenarios. Concretely, we first generate $55$ subdomains and then produce $10$ topics per subdomain, each paired with a persona specification (e.g., role, speaker/addressee, and gender attributes), yielding 55$\times$10=550 topics per country-domain.

\paragraph{Phase 2: English Conversation Creation.} Given these topic specifications, we prompt \texttt{Gemini-2.5 Pro} to generate 2-4 turn English dialogues conditioned on the target country local culture and domain. We constrain the model’s generations to produce spoken dialogues only that are free of personally identifiable information (PII). To reduce lexical leakage from Arabic into the English sources and to encourage semantic (rather than transliteration-based) transfer, we ask the model to use English paraphrases for culturally specific expressions (e.g., ``\textit{God willing}'' rather than the transliterated Arabic ``\textit{inshallah}''). 

We iteratively refine our shared prompt template in pilot runs, using feedback from several team members to improve cultural plausibility, domain coverage, and linguistic naturalness. Applied across all 13 countries, our prompt configuration, instantiated with country- and domain-specific variables, produced 6,050 conversations per country under the 2-4 turns constraint ($\sim$ 3 turns/conversation on average). Examples from Phase 1 and Phase 2 are shown in Figure~\ref{fig:six_country_cards} (Appendix).

\subsubsection{English Source Quality Assurance} \label{app:english_sources_quality}

We screen the generated English conversations using automated checks (format compliance, length bounds, and heuristic PII detection), followed by targeted human review to remove outputs that violate privacy, realism, or guideline constraints.

\begin{figure}[!ht]
    \centering
    \includegraphics[width=0.49\textwidth]{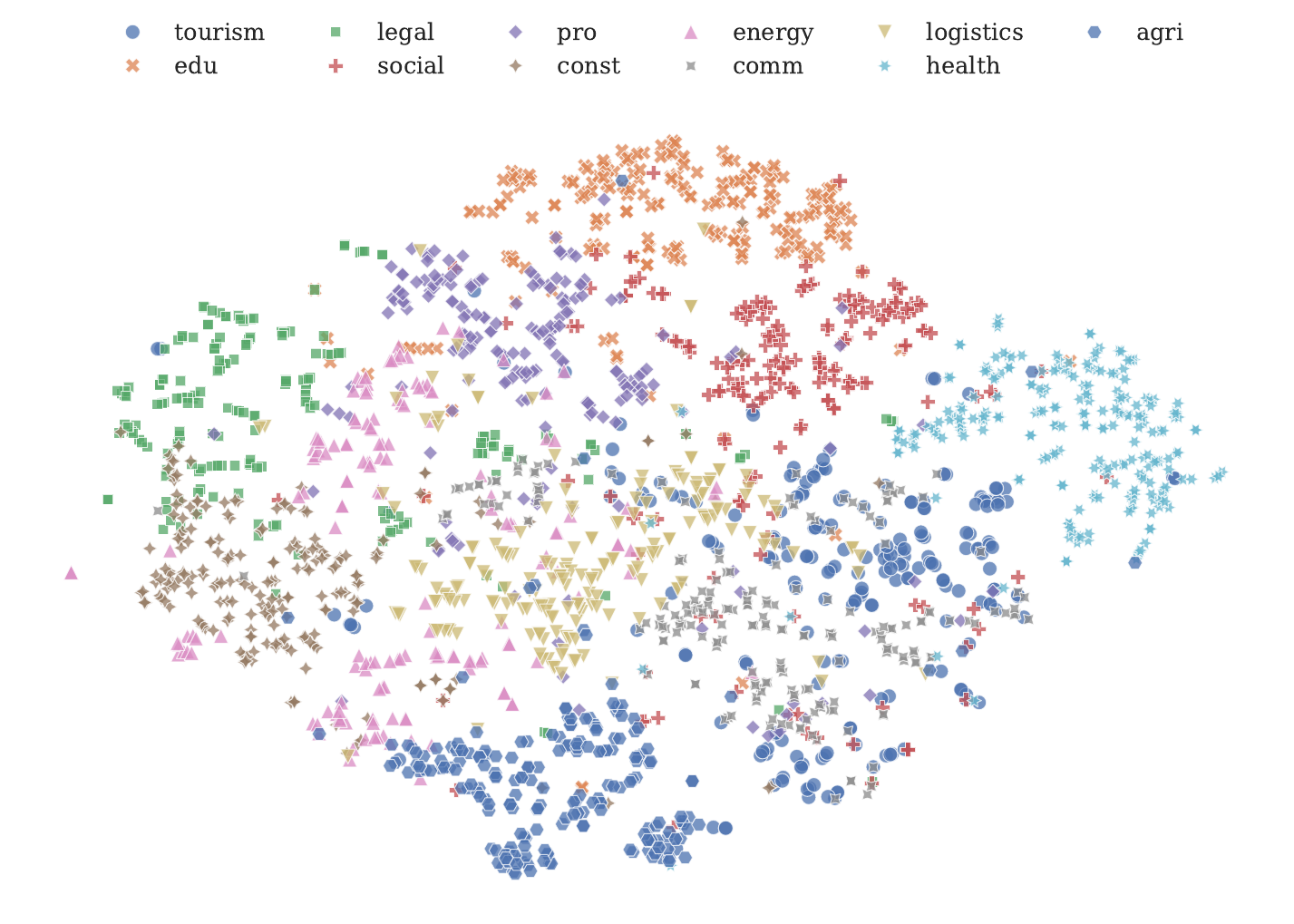}
    \caption{t-SNE projection of generated English conversations, shown for the Moroccan context as a representative example. Conversations are grouped by domain using source embeddings from our two-phase pipeline; these trends are consistent across the other countries.}
    \label{fig:gemini_english_tsne}
\end{figure}

\paragraph{Diversity and Redundancy.} We assess lexical variety using the proportion of unique bigrams (unique/total) computed per domain, identifying the ratio to range from 0.47 to 0.62. To estimate semantic redundancy, we embed each conversation (mean pooled sentence embeddings)\footnote{The model used is \texttt{all-MiniLM-L6-v2}, Available at \url{https://huggingface.co/sentence-transformers/all-MiniLM-L6-v2}.}, and compute cosine similarity over all English conversation pairs; the mean similarity is 0.20, suggesting limited near-duplication/semantic and topical diversity. Figure~\ref{fig:gemini_english_tsne} visualizes the diversity of the generated English sources using t-SNE.

\paragraph{Linguistic and Cultural Screening.} Because LLM-generated sources can contain artifacts (e.g., unnatural phrasing, implausible cultural details, or mismatches with persona/gender specifications), participants were instructed to audit each source conversation before translation and to flag and discard items that violated the guidelines. On average, $2.94$\% of the sentences were skipped by the participants and marked as irrelevant.

\paragraph{English Sources Quality Check} \label{app:english_source_quality_check}
To verify the quality of the English source conversations prior to translation, we incorporated a human validation step into the translators’ workflow. Specifically, we added a dedicated “Comments” column in each participant’s assigned spreadsheet so that translators could flag problematic turns and provide concrete feedback. We instructed participants to skip any conversation that \textit{(i)} did not align with the target community’s cultural context or \textit{(ii)} included culture-specific references that were incorrect for that country/dialect.
This process surfaced several culture-mismatch cases. For instance, in the Jordanian track, an annotator flagged an expression that is common in Syrian usage but not in Jordanian Arabic (e.g., “green light, dead”) and noted that the surrounding scenario relied on assumptions that do not reflect the local context. In the Saudi track, a translator identified a factual inconsistency where the conversation described performing Umrah in Medina rather than in Mecca. Such cases were excluded from the translation set to prevent downstream evaluation from being confounded by culturally inaccurate or factually incorrect source content.

\subsection{Human Dialectal Arabic Translation}
Once the English conversations were generated, we distributed them to the corresponding country teams. Each conversation was translated by a primary translator, who produced a single dialectal Arabic translation for each turn. To promote quality and consistency across contributors and dialects, we implemented the following procedures:

\paragraph{Participant Selection and Training.} We recruited translators who \textit{(i)} self-identified as primary speakers of the target local dialect and \textit{(ii)} reported advanced proficiency in English. The entire cohort was drawn  from academic settings (primarily MA/MSc and PhD students across disciplines). We conducted an initial live training session covering task requirements and guideline conventions; the recording and comprehensive written guidelines were shared for later reference. After one to two weeks, we provided targeted feedback based on sampled translations, focusing on recurrent issues. Throughout the process, country leads conducted ongoing quality control by reviewing submitted translations, providing corrective guidance, and coordinating revisions when needed. Leads also reported periodic progress updates. 

\paragraph{Translation Guidelines.} We provided detailed translation guidelines instructing participants to render each turn in their local Arabic dialect while preserving the meaning of the English source. Contributors were asked to use Arabic script and to avoid rewriting turns into formal MSA; instead, they were encouraged to use colloquial phrasing typical of their variety, without enforcing a single standardized orthography. Translations were produced at the turn level, with instructions to \textit{(i)} maintain \textbf{semantic faithfulness}, \textit{(ii)} adhere to the assigned \textbf{persona attributes} (e.g., role/occupation) and \textbf{speaker-addressee gender configuration}; and \textit{(iii)} follow an appropriate \textbf{social register}. We also provided guidance on  \textit{code-switching:} participants could include commonly used borrowed terms (rendered in Latin script) when they are conventional in the target community and lack a natural dialectal alternative, including terms in English or other locally prevalent contact languages such as French or Spanish. While manual translation was prioritized, we permitted the use of AI assistance under narrowly defined conditions, accompanied by rigorous human post-editing to ensure correctness and dialectal authenticity (see Appendix~\ref{app:ai_use_translation} for analysis of tool usage).


The full translation guidelines are available in the repository mentioned previously. We additionally conducted a post-task survey to document translation challenges and contributor feedback (Appendix \ref{app:translation_challenges}).

\subsection{Translation Revision}

\paragraph{Revision Guidelines.} To improve data quality, we conducted a peer review and revision phase. Each translated conversation was assigned to a second participant from the same country for cross-evaluation. Reviewers assessed each turn along six dimensions: \textit{dialectal authenticity}, \textit{speaker-addressee gender alignment}, \textit{register appropriateness}, \textit{semantic faithfulness}, \textit{punctuation}, and \textit{code-switching} consistency.\footnote{We provided a rubric with examples to guide contributors in this process.} Reviewers then issued an overall decision: \textit{Accept}, \textit{Minor edit} (mechanical corrections such as punctuation/typos), or \textit{Major issue} (substantive problems affecting meaning, register, or metadata alignment). When a translation used a regional dialect different from the reviewer's own, the rubric restricted edits to mechanical corrections only; reviewers were instructed not to alter the dialect-specific  phrasing or vocabulary. Items flagged with major issues were escalated for follow-up (revision by the original translator and/or adjudication by the country lead). Reviewers assigned a difficulty score used for test-set selection; specifically, this metric was employed to penalize simplistic and trivial turns such as ``thank you,'' thereby ensuring a more rigorous evaluation set. Country leads closely monitored progress and resolved escalations. The full revision guidelines are available in the repository mentioned previously.

\paragraph{Revision Insights.} In the revision phase, which was strictly human-only, $68.4$\% of turns remained unchanged, $30.6$\% required minor edits, and 1\% were flagged for major issues. For turns that were edited, the mean normalized edit distance (turn-level Levenshtein distance divided by character count) was $16.9$\%. Beyond structural edits, we assessed the quality of the final output across three dimensions: \textit{dialectal authenticity} (9.03/10), \textit{register appropriateness} (9.40/10), and \textit{semantic faithfulness} (9.36/10).  These high scores, averaged across all target countries, demonstrate that the final output met strict standards of native-speaker authenticity. Table \ref{tab:revision_comparison} (Appendix) illustrates examples of corrections made during the revision phase.

To assess revision reliability, we measured inter-rater agreement on an overlapping subset in which two reviewers from the same country independently reviewed $50$-$110$ shared turns. We report agreement on the three-way decision label (Accept/Minor/Major): the mean exact match rate is $68.2$\%, and the mean  Gwet's AC1 score is $0.65$\footnote{We report Gwet's AC1~\cite{000711006X126600} instead of Cohen's Kappa due to the high class imbalance in our data (a high prevalence of ``Accept'' decisions). In such distributions, Cohen's Kappa penalizes high agreement (the ``Kappa Paradox''), whereas Gwet's AC1 provides a more robust estimate of chance agreement.}.


\paragraph{Preprocessing and Normalization.} We applied a three-step cleaning procedure to the revised turns: \textit{(i)} NFKC Unicode normalization, \textit{(ii)} punctuation normalization, and \textit{(iii)} whitespace cleaning.

Final post-revised examples of the Alexandria parallel conversations are presented in Table \ref{tab:alexandria_qualitative_final} (Appendix).

\subsection{Project Management and Feedback Loops}\label{app:project_management}
To support data quality and maintain steady progress over the project lifecycle, we used a structured coordination process. We held weekly meetings with all participants to review progress and surface recurring issues; feedback from these meetings was used to iteratively refine both the translation guidelines and the annotation platform. Day-to-day coordination occurred through a dedicated Slack workspace, complemented by bi-weekly reminders to keep the workflow on track. Additionally, country leads met every 3–4 weeks to review team-specific progress, address bottlenecks, and consolidate high-level observations and recommendations for subsequent iterations.

\subsection{Alexandria Characteristics}

\paragraph{Dataset Statistics.} The final dataset comprises $34,488$ multi-turn conversations, totaling $107$K turns. Table \ref{tab:revision_dataset_stats} summarizes Alexandria after the revision phase, broken down by country and domain. 
Dataset size varies by country, largely reflecting differences in contributor availability across country teams. 
Despite these differences, each country covers $11$ domains (i.e., no domain is missing for any country/dialect group). On average, a dialectal turn contains $13.23$ words. Table~\ref{tab:alexandria_comparison} compares Alexandria to prior resources along size, domain coverage, and available annotations.
\begin{table*}[t]
\centering
\sisetup{
    group-separator={,},
    table-format=5.0
}
\resizebox{\textwidth}{!}{%
\begin{tabular}{
    l@{\hskip 0.5em}  
    @{\hskip 1em}S S S S      
    @{\hskip 0.8em}S S S      
    @{\hskip 0.8em}S S        
    @{\hskip 0.8em}S S S S    
} 
\toprule
\textbf{Domain} & 
\multicolumn{4}{c}{\textbf{Levant}} & 
\multicolumn{3}{c}{\textbf{Gulf}} & 
\multicolumn{2}{c}{\textbf{Nile}} & 
\multicolumn{4}{c}{\textbf{Maghreb}} \\ 
\cmidrule(lr){2-5} \cmidrule(lr){6-8} \cmidrule(lr){9-10} \cmidrule(lr){11-14}

& {JO} & {LB} & {PS} & {SY} & 
  {SA} & {OM} & {YE} & 
  {EG} & {SD} & 
  {LY} & {MA} & {MR} & {TN} \\ 
\midrule

\rowcolor{rowgray}
\dicon{\faTractor} Agriculture/Farming & 825 & 1140 & 1770 & 931 & 1162 & 915 & 529 & 583 & 163 & 231 & 570 & 970 & 481 \\
\dicon{\faCreditCard} Commerce/Transactions & 750 & 1004 & 1595 & 749 & 1020 & 650 & 579 & 506 & 201 & 160 & 445 & 757 & 401 \\
\rowcolor{rowgray}
\dicon{\faBuilding} Construction/Real Estate & 859 & 995 & 1761 & 861 & 1161 & 974 & 696 & 660 & 225 & 271 & 574 & 673 & 485 \\
\dicon{\faGraduationCap} Education/Academia & 816 & 1191 & 1513 & 831 & 1017 & 1079 & 563 & 549 & 170 & 220 & 601 & 863 & 551 \\
\rowcolor{rowgray}
\dicon{\faLightbulb} Energy/Resources & 786 & 1048 & 1715 & 928 & 1177 & 937 & 587 & 625 & 189 & 243 & 447 & 719 & 470 \\
\dicon{\faComments} Everyday/Social & 967 & 1215 & 1697 & 787 & 1020 & 888 & 642 & 604 & 175 & 210 & 595 & 824 & 550 \\
\rowcolor{rowgray}
\dicon{\faHeartbeat} Healthcare/Medical & 727 & 1240 & 1728 & 781 & 1043 & 895 & 548 & 487 & 164 & 253 & 556 & 948 & 522 \\
\dicon{\faGavel} Legal/Financial & 693 & 1006 & 1566 & 757 & 857 & 753 & 496 & 539 & 177 & 174 & 481 & 642 & 412 \\
\rowcolor{rowgray}
\dicon{\faTruck} Logistics/Transport & 842 & 1020 & 1512 & 950 & 1234 & 842 & 629 & 646 & 189 & 187 & 593 & 877 & 515 \\
\dicon{\faBriefcase} Professional/Workplace & 845 & 1220 & 1810 & 959 & 1112 & 866 & 549 & 645 & 178 & 253 & 480 & 709 & 526 \\
\rowcolor{rowgray}
\dicon{\faSuitcaseRolling} Tourism/Hospitality & 720 & 1161 & 1596 & 884 & 1004 & 815 & 608 & 608 & 190 & 216 & 567 & 878 & 460 \\ 

\midrule[\heavyrulewidth]
\textbf{\dicon{\faChartBar} Total} & \textbf{8,830} & \textbf{12,240} & \textbf{18,263} & \textbf{9,418} & \textbf{11,807} & \textbf{9,614} & \textbf{6,426} & \textbf{6,452} & \textbf{2,021} & \textbf{2,418} & \textbf{5,909} & \textbf{8,860} & \textbf{5,373} \\ 
\bottomrule
\end{tabular}%
}
\caption{Post-revision turn statistics in the Alexandria dataset across domains and dialects.}
\label{tab:revision_dataset_stats}
\end{table*}

\begin{figure}[!ht]
    \centering
    \includegraphics[width=0.49\textwidth]{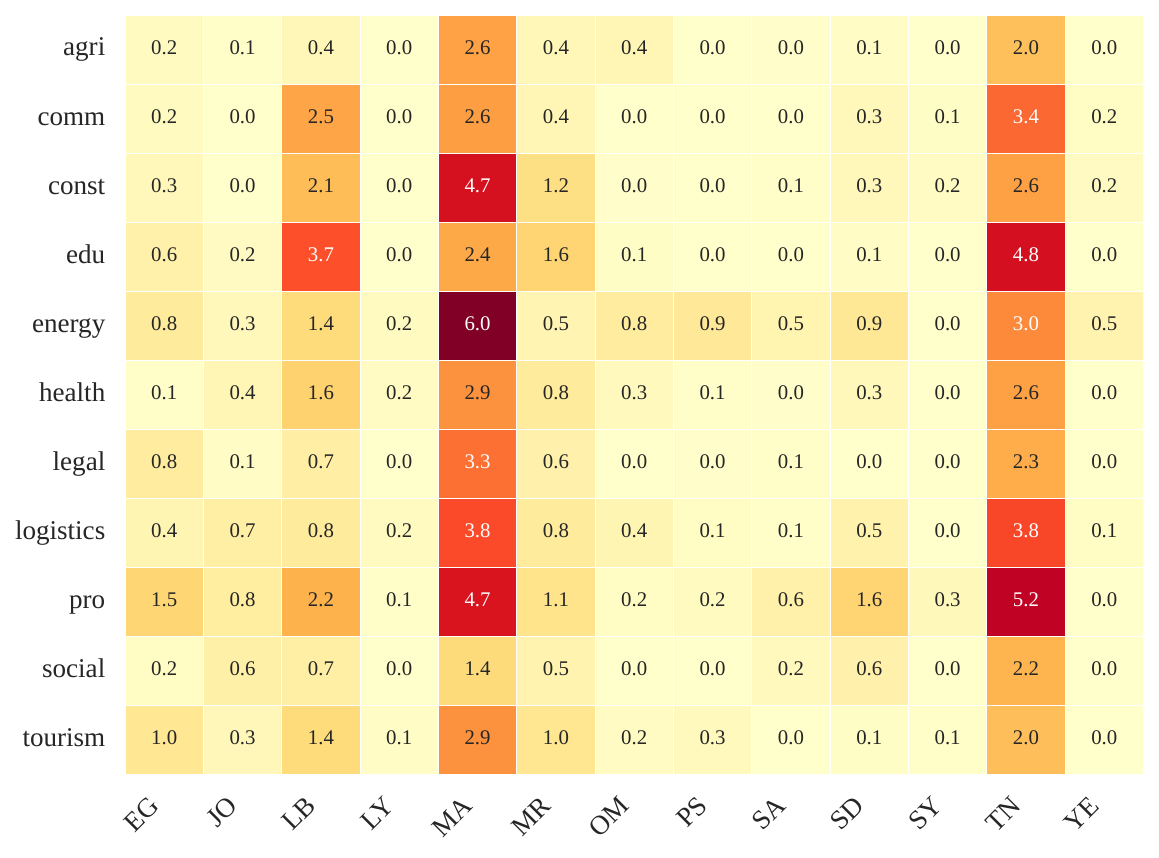}
    \caption{Average Code-Mixing Index across dialects and domains in Alexandria dataset.}
    \label{fig:cmi_heatmap}
\end{figure}

\paragraph{Code-Switching Rates.} Figure~\ref{fig:cmi_heatmap} reports the Code-Mixing Index~\cite{das-gamback-2014-identifying} (Latin vs Arabic script) by dialect group and domain.  Moroccan and Tunisian varieties show consistently higher code-mixing, Lebanon exhibits moderate levels (notably  in education and communication), and most other dialect groups remain low.

\paragraph{Gender Direction Distribution.} The final version of Alexandria includes four speaker-addressee gender configurations: F$\rightarrow$M (33.19\%), M$\rightarrow$F (32.78\%), M$\rightarrow$M (21.43\%), and F$\rightarrow$F (12.60\%).

\begin{figure*}[!ht]
    \centering
    \includegraphics[width=\textwidth]{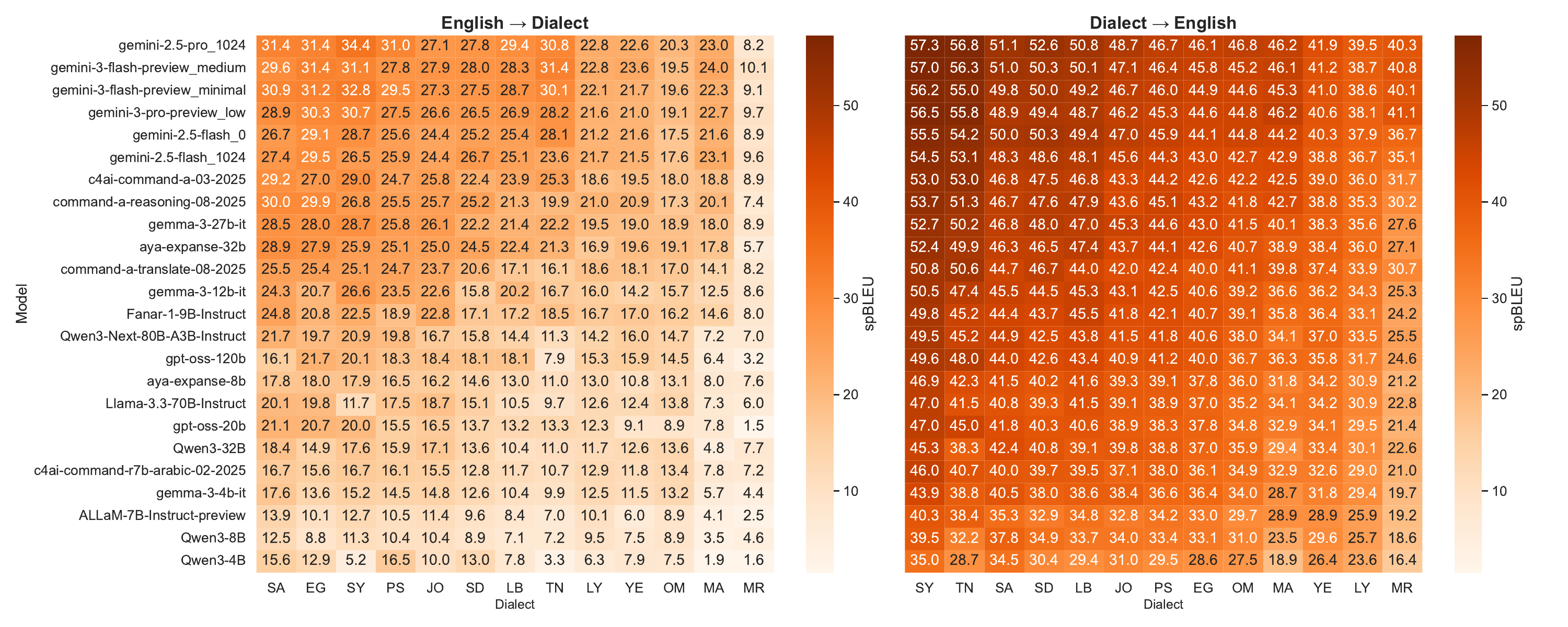}
    \caption{Context-aware MT performance (spBLEU) across 13 dialects. Results reveal a significant directional asymmetry: models perform consistently stronger on Dialect $\to$ English (right) than English $\to$ Dialect (left). Maghrebi dialects (e.g., MR, MA, TN) remain the most challenging across all models.}
    \label{fig:scores_spbleu_heatmap}
\end{figure*}

\paragraph{Data Splits and Release.} The Alexandria dataset is partitioned into four splits: training, public development, public test, and private test.\footnote{A private test set is withheld to facilitate future open evaluations (e.g., leaderboards, shared tasks).} To ensure balanced representation, the public development and test sets are stratified equally across dialect groups, genders, and translators. Specifically, each country-domain pair contributes $\sim$100 turns ($\sim$30 conversations) to the public test and $\sim$ 50 turns to public development, yielding $\sim$1,100 test turns and $\sim$550 dev turns per country (11 domains x 100/50). The remaining data are allocated to the training and private test sets.

\section{Evaluation}
\subsection{Evaluation Setup}
We use Alexandria public test set to evaluate English$\leftrightarrow$Arabic translation across a diverse set of API access (closed-weight) and open-weight Arabic-capable LLMs. Exploiting Alexandria's conversational structure and metadata (persona role and speaker$\rightarrow$addressee gender configuration), we evaluate three input settings: \textit{(i) Turn-level}, translating a single turn in isolation; \textit{(ii) Context-level}, translating a turn given the preceding source-side dialogue turns; and \textit{(iii) Conversation-level}, translating the entire conversation in a single pass with explicit turn delimiters. In all settings, we prepend the relevant metadata to the input. Figure \ref{fig:llm_evaluation_prompts} (Appendix) provides the prompt templates used for generation.

We evaluate $24$ Arabic-capable LLMs including the  \texttt{Gemini}, \texttt{Qwen $3$}, \texttt{Gemma}, and \texttt{Command A} (Table~\ref{tab:evaluated_llms}, Appendix), including both standard and "reasoning" variants. Unless otherwise noted, decoding uses greedy generations (temperature=0).

\subsection{Automatic Evaluation}
We report reference-based surface metrics using SacreBLEU: spBLEU~\cite{goyal-etal-2022-flores} (SentencePiece with the flores200 tokenizer), and chrF++~\cite{popovic-2017-chrf} (robust for rich morphology). We avoid model-based metrics such as COMET~\cite{rei-etal-2020-comet} due to the limited reliability of dialectal Arabic. 

\subsection{Human Evaluation}
For human evaluation, we focus on English$\rightarrow$dialect, which is more sensitive to dialectness and lexical homogenization (i.e., MSA leakage).  Native speakers evaluate outputs from six selected LLMs\footnote{These models were selected based on the diversity of their automatic evaluation scores.} on \textit{(i) semantic adequacy} (5-point Crosslingual Semantic Text Similarity [XSTS] scale~\cite{agirre-etal-2012-semeval}), to measure meaning preservation regardless of variety; \textit{(ii) gender accuracy} (Pass/Fail/NA); and \textit{(iii) dialectness \& Fluency} (1--5). 
Further details on the scoring rubrics are provided in Appendix~\ref{app:human_eval}.

We selected 1--3 evaluators per country.\footnote{EG, JO, SD, TN, and YE each had one evaluator.} Each evaluator rated $\sim$500 items, stratified across models and domains. For countries with $\geq$2 evaluators, evaluator pairs additionally rated an overlapping subset of $\sim$50 items to estimate consistency. 
We computed inter-rater agreement for three criteria: gender accuracy achieved a mean Gwet’s AC1 of $0.970$ (averaged across countries), while semantic adequacy and dialectness/fluency yielded Intraclass Correlations (ICC(2,k)) of $0.45$ and $0.56$, respectively, indicating fair-to-moderate agreement.

\begin{figure*}[!ht]
    \centering
    \includegraphics[width=\textwidth]{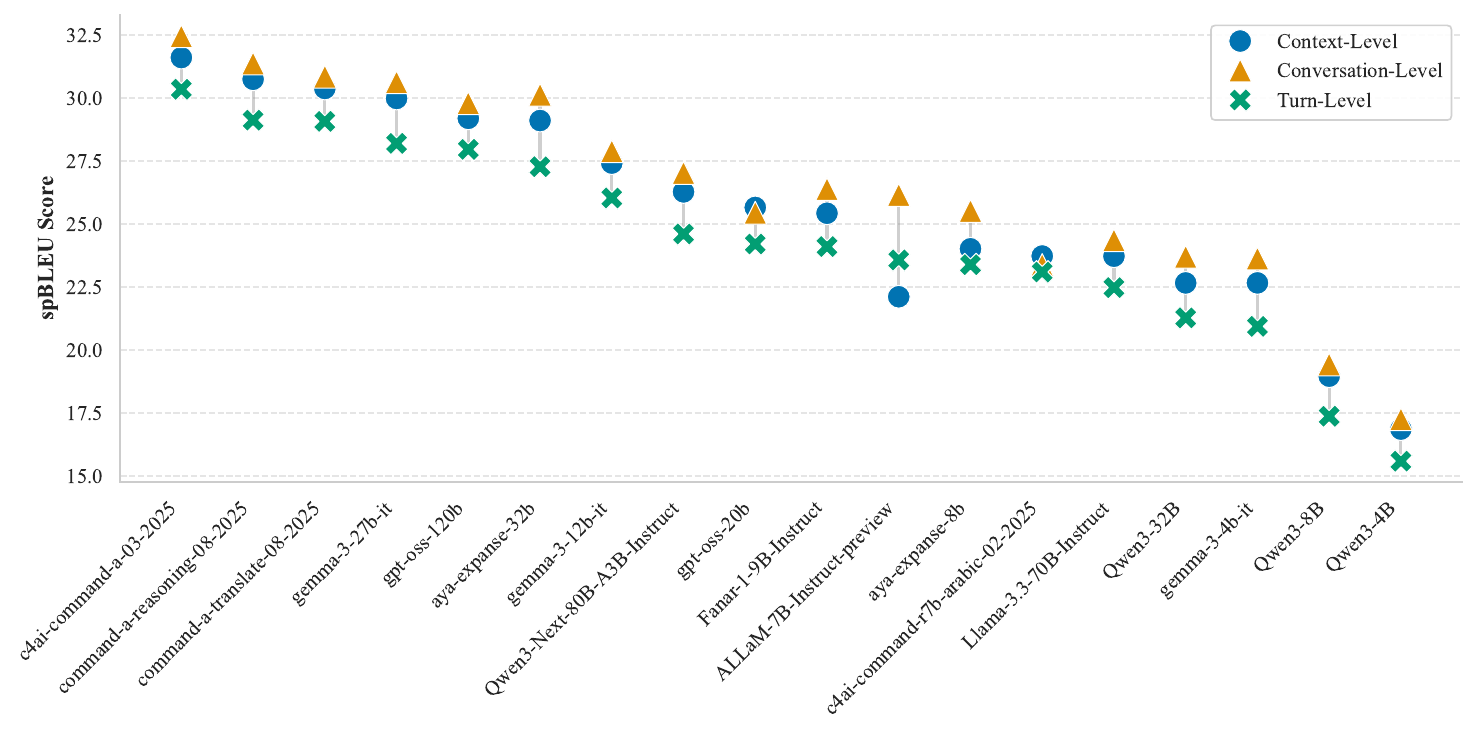}
    \caption{Comparison of evaluation scenarios across various models. The reported spBLEU scores represent an average across all dialects for both translation directions.}
    \label{fig:evals_scenarios}
\end{figure*}

\section{Results and Discussion}
We focus on the context-level conversation setting, which better matches conversational MT: the system translates the current turn given only the preceding dialogue history, without access to future turns (Figure \ref{fig:evals_scenarios} presents comparison against the other settings). While conversation-level translation (full-conversation input)  yields higher raw scores, it represents a more permissive, offline setting. Because spBLEU and chrF++ are highly correlated in our experiments (Pearson $r=0.9079$), we report spBLEU in the main text for compactness and include chrF++ in Appendix \ref{app:automatic_results}.

\subsection{Automatic Evaluation Results}

\paragraph{Per-Dialect Results.} Figure \ref{fig:scores_spbleu_heatmap} reports context-level spBLEU across the $13$ country-level dialect groups, with scores averaged over city-level varieties and domains. We observe a clear directional asymmetry: dialect$\rightarrow$English achieves consistently higher scores than English$\rightarrow$dialect. Among the evaluated models, the Gemini variants (specifically \texttt{Gemini-2.5-pro} and \texttt{Gemini-3-flash}) achieve the strongest performance across both directions. Performance also varies substantially by dialect group: models tend to perform best on Egyptian and Levantine varieties (e.g., SY, LB, JO), possibly due to training data availability for these varieties, while Maghrebi varieties pose a greater challenge, with Mauritanian yielding the lowest scores.

\begin{figure}[!ht]
    \centering
    \includegraphics[width=0.49\textwidth]{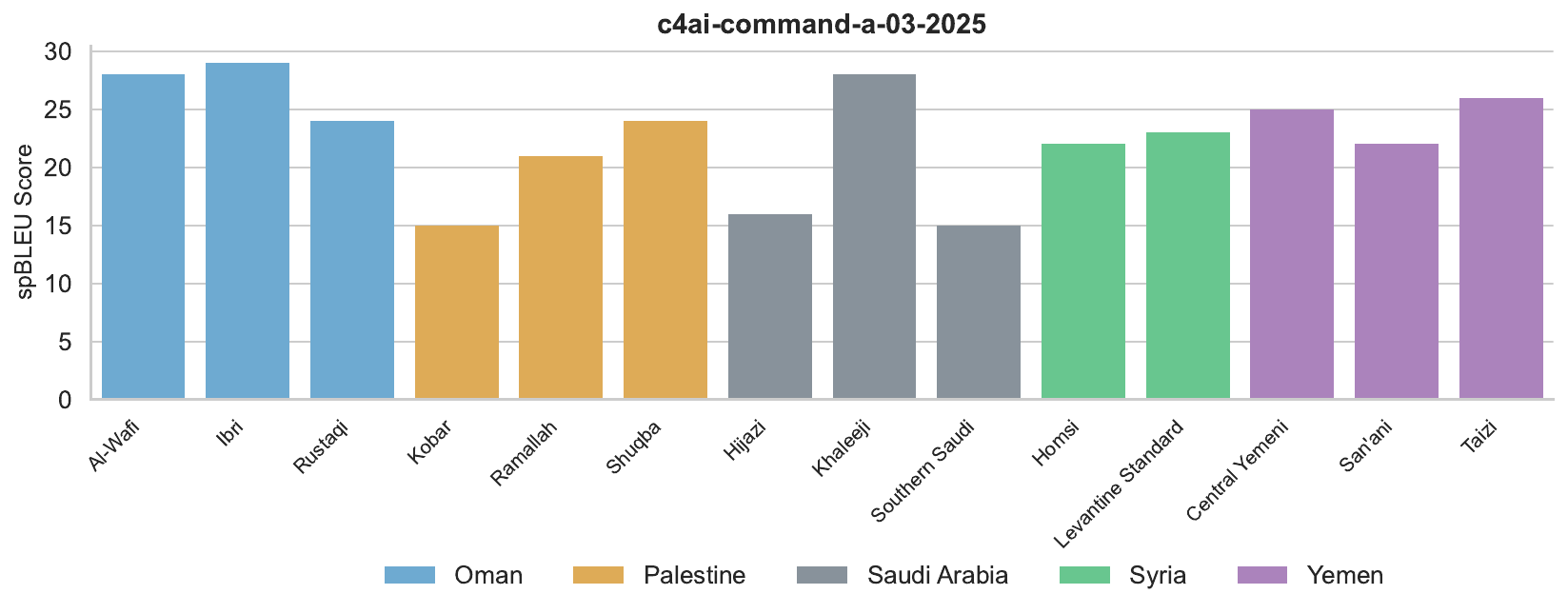}
    \caption{Intra-country performance variance (English $\to$ Sub-Dialect). Scores for selected sub-dialects reveal systematic difficulty gaps within countries (e.g., urban vs. rural Palestinian varieties), with consistent model rankings across sub-dialects.}
    \label{fig:subdialects_scores}
\end{figure}

\paragraph{Per-Sub-Dialect Results.} We further evaluate performance at the sub-dialect (city-level) granularity. We select the three best-performing LLMs from the previous section (based on context-level spBLEU averaged across dialect groups and domains) and evaluate them on each sub-dialect; Figure~\ref{fig:subdialects_scores} summarizes the results. Focusing on \emph{within-country} variation, we observe that sub-dialect rankings are broadly consistent across models: while absolute scores differ, the relative ordering of sub-dialects within a country is largely stable, suggesting systematic sub-dialect difficulty that generalizes across model families. Results from additional models are provided in Figure~\ref{fig:subdialects_scores_app} (Appendix). 


\begin{figure}[!ht]
    \centering
    \includegraphics[width=0.49\textwidth]{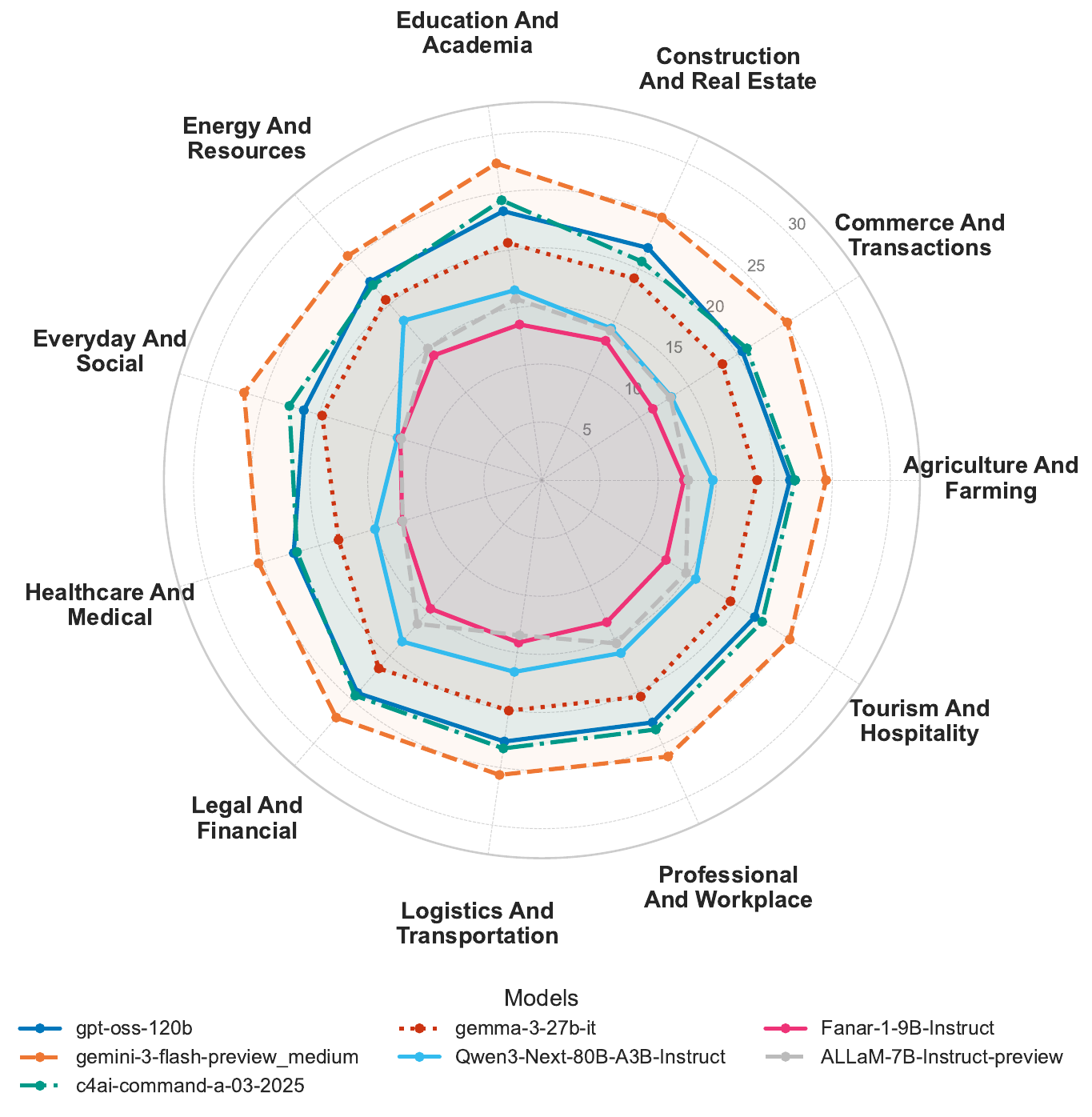}
    \caption{Domain robustness analysis (English $\to$ Dialect). The plot illustrates  average per-domain spBLEU scores for a subset of models across all 11 domains.}
    \label{fig:domain_to_ar_robust}
\end{figure}

\paragraph{Per-Domain Results.} Figure \ref{fig:domain_to_ar_robust} presents spBLEU for a subset of LLMs across the 11 domains, averaged across countries (English$\to$dialect). Model rankings are highly stable across domains: the same top-tier models (e.g., \texttt{Gemini-3-flash} and \texttt{Command A}, consistently achieve the highest scores, while smaller open-weight models (e.g.,  \texttt{ALLaM-7B} and \texttt{Fanar-1-9B}) remain in the lower tier. The limited crossing of model performance curves across domains suggests that, under this evaluation setup, overall model strength is a strong predictor of model performance across domains, with comparatively little evidence of domain-specific specialization. dialect$\to$English results are reported in Figure~\ref{fig:domain_to_en_robust} (Appendix).

\begin{table*}[htbp]
\centering
\resizebox{\textwidth}{!}{%
\begin{tabular}{llccccccccc}
\toprule
\textbf{Model} & \textbf{Metadata} & \textbf{EG} & \textbf{JO} & \textbf{LB} & \textbf{MA} & \textbf{PS} & \textbf{SA} & \textbf{SY} & \textbf{TN} & \textbf{YE} \\
\midrule
\textbf{NLLB-200-3.3B} & \textbf{N/A} & 17.16 & 18.44 & 17.06 & 9.82 & 18.57 & 17.96 & 22.24 & 10.64 & 11.17 \\
\midrule
\textbf{gemma-3-12b-it} & \textbf{None} & \underline{25.54} & 23.54 & 20.82 & 11.33 & \underline{24.12} & 25.65 & \underline{25.79} & \underline{16.84} & 17.39 \\
 & \textbf{Partial} & 25.22 & \underline{23.77} & \textbf{\underline{21.30}} & \underline{11.47} & 23.19 & \underline{25.97} & \underline{25.79} & 16.30 & 16.91 \\
 & \textbf{Full} & 25.11 & 23.14 & 20.96 & 11.34 & 23.39 & 24.39 & 24.90 & 12.00 & \underline{17.50} \\
\midrule
\textbf{aya-expanse-32b} & \textbf{None} & 25.06 & 22.26 & 19.77 & \underline{11.52} & 21.31 & 23.54 & \underline{26.31} & \underline{15.97} & 14.01 \\
 & \textbf{Partial} & \underline{25.32} & \underline{22.66} & \underline{20.49} & 10.96 & 21.84 & \underline{23.82} & 26.25 & 15.91 & \underline{14.09} \\
 & \textbf{Full} & 24.35 & 21.52 & 14.51 & 7.54 & \underline{21.90} & 23.27 & 24.68 & 8.34 & 12.99 \\
\midrule
\textbf{c4ai-command-a-03-2025} & \textbf{None} & 28.78 & 24.27 & 20.20 & 18.60 & \textbf{\underline{25.34}} & 28.88 & \textbf{\underline{27.74}} & \textbf{\underline{23.91}} & 18.58 \\
 & \textbf{Partial} & 29.07 & 24.31 & 20.88 & 19.19 & 25.06 & 28.98 & 27.70 & 20.36 & 19.34 \\
 & \textbf{Full} & \textbf{\underline{29.45}} & \textbf{\underline{25.41}} & \underline{21.13} & \textbf{\underline{20.01}} & 25.29 & \textbf{\underline{29.40}} & 26.96 & 20.79 & \textbf{\underline{20.51}} \\
\bottomrule
\end{tabular}}
\caption{Single-turn English-to-Dialect spBLEU scores across nine Arabic dialects. LLMs are evaluated under metadata ablations: \textit{None}, \textit{Partial} (gender only), and \textit{Full} (complete metadata). \textbf{Bold} denotes the overall column maximum; \underline{underline} denotes the maximum per model.}
\label{tab:metadata_nllb}
\end{table*}

\paragraph{Impact of prompt metadata and comparison against NLLB.} As most of our experiments rely on conditioned prompts with metadata, we conducted an ablation study to quantify the performance gains derived from these contextual cues and to anchor our LLM evaluations against a traditional machine translation baseline. Specifically, we evaluated a subset of LLMs at the single-turn level under three distinct metadata conditions: \textit{None} (no metadata provided), \textit{Partial} (participant gender only), and \textit{Full} (complete metadata). We then compared these outcomes against NLLB-200-3.3B~\cite{nllbteam2022languageleftbehindscaling}, which natively supports 9 of the 13 Arabic dialects present in Alexandria.
As detailed in Table~\ref{tab:metadata_nllb}, the evaluated LLMs consistently outperform the NLLB-200-3.3B baseline across all supported dialects, even in the unconditioned \textit{(None)} setting. However, the metadata ablation results are mixed rather than uniformly monotonic. While metadata can improve performance in some cases (e.g. c4ai-command-a-03-2025), these gains are not consistent across models or dialects. For gemma-3-12b-it, the differences across None, Partial, and Full are generally small and sometimes negative, and for aya-expanse-32b the Full setting is often substantially worse than either None or Partial. Overall, these results suggest that contextual metadata can be helpful, but its effect is model and dialect dependent rather than a general trend of incremental improvement.

\begin{table}[htbp]
\centering
\resizebox{0.48\textwidth}{!}{%
\begin{tabular}{lcc}
\toprule
\textbf{Country} & \textbf{Lexical Overlap} & \textbf{Correlation with spBLEU} \\
\midrule
\textbf{EG} & 0.17 & 0.3970 \\
\textbf{JO} & 0.21 & 0.4335 \\
\textbf{LB} & 0.15 & 0.3967 \\
\textbf{MA} & 0.10 & 0.2731 \\
\textbf{PS} & 0.19 & 0.3943 \\
\textbf{SA} & 0.23 & 0.4824 \\
\textbf{SY} & 0.22 & 0.1911 \\
\textbf{TN} & 0.14 & 0.3606 \\
\textbf{YE} & 0.19 & 0.4414 \\
\bottomrule
\end{tabular}}
\caption{Lexical overlap between dialectal references and NLLB-generated MSA translations, alongside their correlation with spBLEU scores.}
\label{tab:lexical_overlap}
\end{table}
\begin{table}[htbp]
\centering
\resizebox{0.48\textwidth}{!}{%
\begin{tabular}{lccc}
\toprule
\textbf{Country} & \textbf{Avg spBLEU (No Latin)} & \textbf{Avg spBLEU (Has Latin)} & \textbf{CMI Correlation} \\
\midrule
\textbf{EG} & 27.86 & 21.86 & -0.13 \\
\textbf{JO} & 24.17 & 18.85 & -0.06 \\
\textbf{LB} & 22.30 & 17.13 & -0.14 \\
\textbf{MA} & 20.81 & 16.33 & -0.14 \\
\textbf{PS} & 23.75 & 17.56 & -0.05 \\
\textbf{SA} & 27.27 & 28.28 & 0.02 \\
\textbf{SY} & 26.51 & 29.89 & 0.02 \\
\textbf{TN} & 24.41 & 19.39 & -0.21 \\
\textbf{YE} & 19.78 & 21.78 & 0.01 \\
\bottomrule
\end{tabular}}
\caption{Impact of code-switching on translation quality, comparing average spBLEU scores for sentences with and without Latin characters, along with the Code-Mixing Index (CMI) correlation.}
\label{tab:code_switching_impact}
\end{table}

\paragraph{Impact of linguistic distance and code-switching on translation performance.} 
To better understand the variance in translation quality across different regions, we investigated two primary linguistic factors: a dialect's distance from MSA and the prevalence of intra-sentential code-switching. First, we measured the lexical overlap between our dialectal reference texts and MSA translations generated by NLLB. As detailed in Table~\ref{tab:lexical_overlap}, we observed a moderate positive correlation between MSA lexical overlap and spBLEU scores (from English-to-Dialect using c4ai-command-a-03-2025 experiments) across the dataset (e.g., reaching 0.48 for Saudi and 0.44 for Yemeni). This indicates that models consistently yield higher quality translations for sentences—and by extension, dialects like Levantine and Gulf—that share lexical and structural similarities with MSA. Conversely, structurally distant varieties, such as Maghrebi dialects, pose a significantly greater challenge. Second, we analyzed the impact of code-switching, which speakers frequently employ using Latin script (e.g., English or French loanwords) to bridge lexical gaps in technical domains. By comparing sentences with and without Latin characters (Table~\ref{tab:code_switching_impact}), we found that code-mixing degrades translation performance. For the majority of the evaluated dialects (including Egyptian, Jordanian, Lebanese, Moroccan, Palestinian, and Tunisian), average spBLEU scores dropped substantially in the presence of code-switching, a finding further reinforced by a negative correlation with the Code-Mixing Index.

\begin{figure}[!ht]
    \centering
    \includegraphics[width=0.49\textwidth]{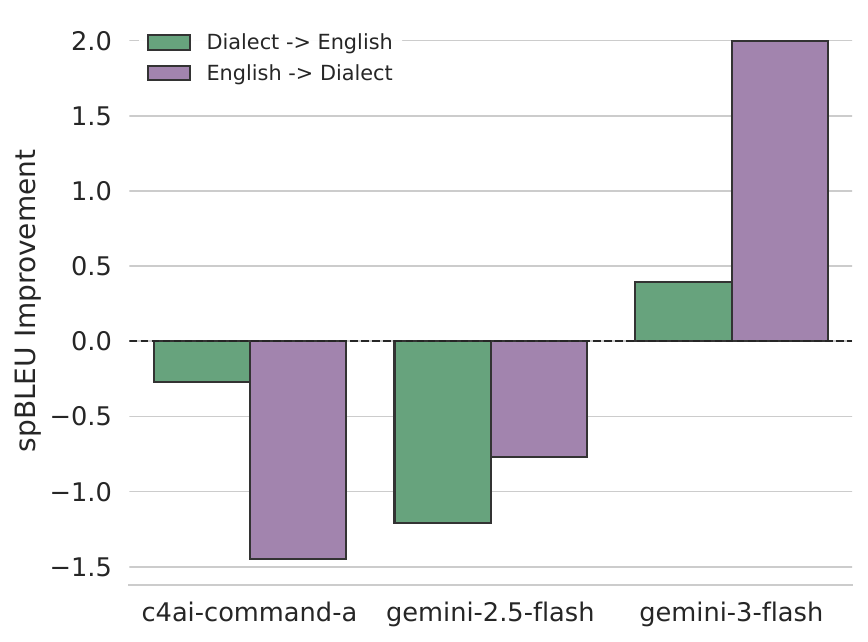}
    \caption{Impact of reasoning on translation performance. The bar chart shows the spBLEU improvement (or degradation) when reasoning is enabled for the evaluated models compared to their non-reasoning baselines.}
    \label{fig:reasoning_impact}
\end{figure}

\paragraph{Does the LLM Thinking help the translation?} Figure \ref{fig:reasoning_impact} presents a comparison between three models using two configurations: one with the thinking process and one without. The results show that the thinking process generally does not help and often hurts translation performance, except for gemini-3-flash. In this case, reasoning boosts average performance by 2.0 spBLEU points for English-to-Dialect and approximately 0.4 points for Dialect-to-English.

\begin{figure}[!ht]
    \centering
    \includegraphics[width=0.49\textwidth]{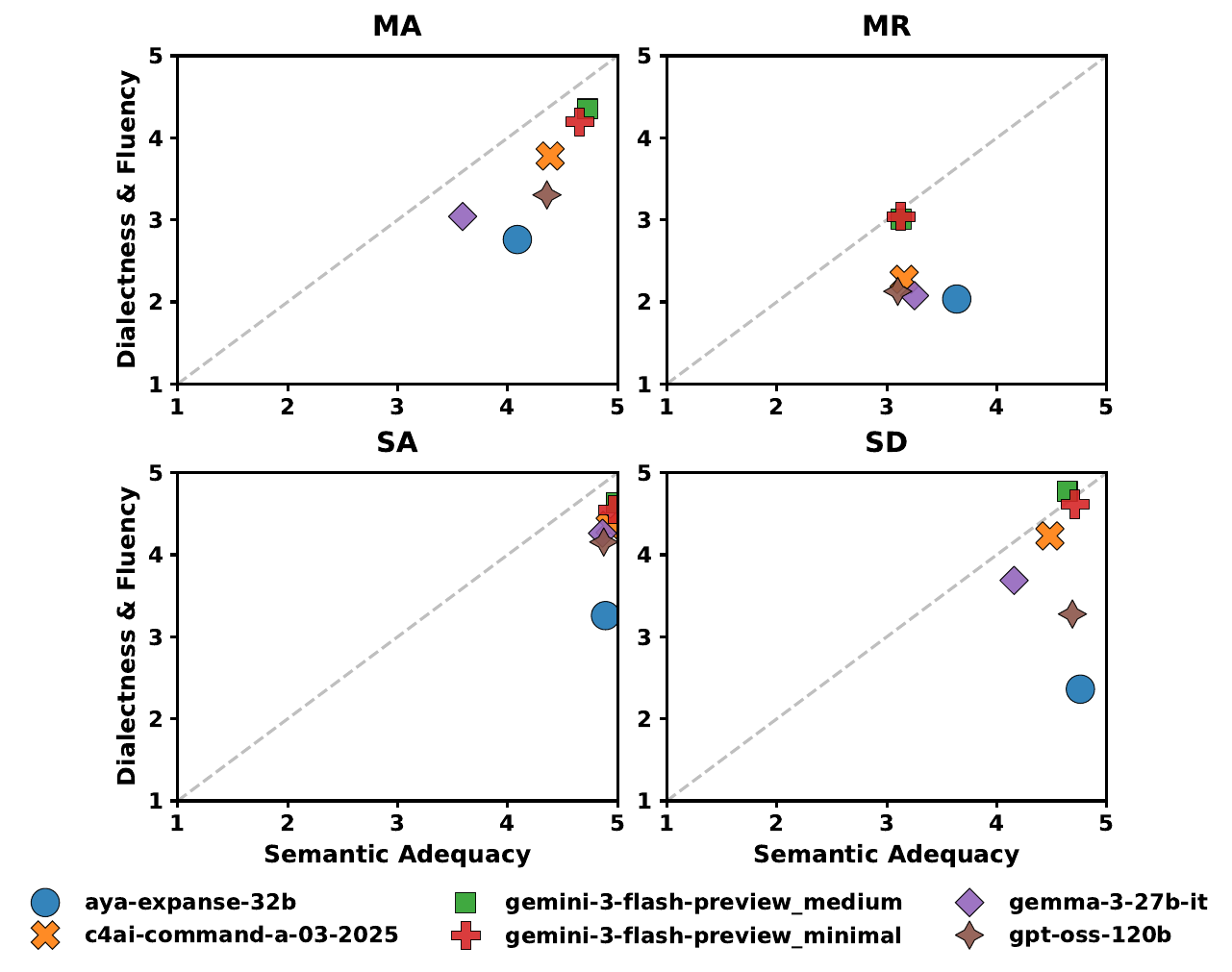}
    \caption{Human evaluation results of Semantic Adequacy vs. Dialectness across four representative dialects.}
    \label{fig:human_main_results}
\end{figure}

\subsection{Human Evaluation Results}
Detailed per-model and per-country human evaluation results are provided in Tables \ref{tab:gender_accuracy_results}, \ref{tab:semantic_adequacy_results}, and \ref{tab:dialectness_fluency_results} (Appendix). Overall, gender accuracy is high (typically >90\%), suggesting that models effectively adhere to explicit gender constraints when provided in the prompt. In contrast, models exhibit significant performance drops in semantic adequacy and, most notably, dialectness/fluency. Across dialects, average semantic adequacy remains above 3/5, whereas dialectness/fluency is substantially lower, dropping to $\sim$2/5 for some model-country pairs. Figure \ref{fig:human_main_results} plots semantic adequacy ($x$-axis) against dialectness/fluency ($y$-axis) for four representative dialect groups: Moroccan, Mauritanian, Saudi, and Sudanese (results for the remaining dialects are reported in Figure \ref{fig:human_app_results}, Appendix). Most data points lie below the diagonal identity line ($y=x$), indicating that models preserve meaning more reliably than they produce dialect-authentic output. We also observe systematic differences across dialects: Saudi and Sudanese varieties tend to achieve higher scores on both axes, while Mauritanian remains the most challenging, with dialectness often near 2.0 even when semantic adequacy exceeds 3.0. Among the evaluated models, \texttt{Gemini-3-flash} and \texttt{Command-A} consistently define the Pareto frontier (offering the strongest adequacy-dialectness trade-off), whereas \texttt{gpt-oss-120b} demonstrates comparatively lower dialectness/fluency across all regions.


\section{Conclusion}

In this work, we introduce Alexandria, a culturally inclusive benchmark covering $13$ Arab countries, designed to evaluate the dialectal capabilities of Arabic-aware LLMs. Curated by a community of $55$ researchers, the dataset comprises $107$K turns ($34,488$ conversations) across $11$ domains. Our evaluations reveal critical gaps in existing models regarding regional dialects, technical terminology, and gender alignment. By releasing Alexandria, we provide a robust framework to address these limitations, fostering the development of more accurate and culturally sensitive language technologies.

\section*{Limitations}

\begin{itemize}
    \item \textbf{Gender Imbalance in Scenarios:} Our dataset exhibits an imbalance in gender transfer directions, specifically a lower frequency of female-to-female interactions ($12.60$\%) compared to other categories. This disparity is an artifact of the Phase 1 generation process; while our prompts explicitly requested diverse personas, the source LLM exhibited a latent bias toward generating mixed-gender or male-dominant scenarios. However, to ensure fair evaluation, the test and development sets were explicitly curated in a balanced manner across gender directions.

    \item \textbf{Technical Lexical Gaps and MSA Leakage:} Annotators reported significant hurdles when translating technical terminology (e.g., in mining, geology, or corporate logistics) that lacks direct dialectal equivalents. In these instances, translators frequently resorted to MSA or code-switching to convey scientific concepts, which may introduce a formal register bias in technical 
    domains.

    \item \textbf{Restricted Closed-Models Evaluation:} We integrated LLMs into our framework following their demonstrated effectiveness in translation tasks, particularly among proprietary architectures. Due to budget constraints, we could not assess the full range of closed-source models and instead focused our evaluation exclusively on the Gemini suite.

    
\end{itemize}

\section*{Ethics Statement}
All Alexandria translations were produced by community participants under a pre-established authorship agreement. Specifically, contributors who translated and revised a minimum of $3,000$ sentences each are included as co-authors to ensure full credit for their substantial labor. Participants who contributed to the project but did not meet this threshold are recognized in the Acknowledgments. To maintain ethical standards, we used English source sentences free of personally identifiable information (PII) and provided participants with rigorous guidelines regarding local norms, data privacy, and informed consent.

\section*{Acknowledgments}

Muhammad Abdul-Mageed acknowledges support from Canada Research Chairs (CRC), the Natural Sciences and Engineering Research Council of Canada (NSERC; RGPIN-2018-04267), the Social Sciences and Humanities Research Council of Canada (SSHRC; 895-2020-1004; 895-2021-1008), Canadian Foundation for Innovation (CFI; 37771), Digital Research Alliance of Canada,\footnote{\href{https://alliancecan.ca}{https://alliancecan.ca}} and UBC Advanced Research Computing-Sockeye.\footnote{\href{https://arc.ubc.ca/ubc-arc-sockeye}{https://arc.ubc.ca/ubc-arc-sockeye}}

In addition to the authors who provided translations for this project, we gratefully acknowledge the help and contributions of the following individuals: 
Mazen Al-Asali, 
Doaa Qawasmeh, 
Vatimetou Mohamed Lemin, 
Abdallah Al-Ameen, 
Muhammed Saeed, 
Hawraa Ramadhan, 
Ro'a Nafi, 
Ghadeer Shalash, 
Saja Ayyad, 
Lina Hamad, 
Asia Albarghouthi, 
Manar Shawahnii, 
Mohammed Anwar Al-Ghrawi, 
Aminetou Yacoub, 
Sumayah Alsakiti, 
Raheeq Mousa, 
Sondos Khieriah, and 
Itidal Fares. Also, we thank Mohammed Akallouch for polishing and enhancing the visual presentation of the paper's main figure.

We also acknowledge support from the Google Cloud Research Credits program (Award GCP19980904), which was utilized during data generation and evaluation API calls.

\bibliography{custom}
\clearpage
\appendix
\appendixpage            
\addappheadtotoc         
\numberwithin{figure}{section}
\numberwithin{table}{section}

\section{Alexandria MT Dataset Creation}
\label{app:data_creation}


\begin{figure*}[p]
    \centering
    \scriptsize 
    \setlength{\lineskip}{2pt}
    
    \noindent
    \begin{minipage}[t]{0.49\linewidth}
        \begin{tcolorbox}[datacard={egyptprimary}{Egypt: Logistics \& Transportation}{\faTruck}]
            \linespread{0.85}\selectfont
            \vspace{1mm}
            \begin{center}
                \activebadge{egyptprimary}{egypthighlight}{\faWarehouse \ Warehouse}
                \passivebadge{\faBox \ Freight}
                \passivebadge{\faShip \ Port}
                \enspace {\tiny \textit{... (+52)}}
            \end{center}
            \vspace{1mm}
            \begin{tcolorbox}[enhanced, colback=egyptbg, colframe=egyptbg, borderline west={2pt}{0pt}{egypthighlight}, arc=1mm, left=1.5mm, top=0.5mm, bottom=0.5mm]
                {\tiny \textbf{\color{egypthighlight} \faListUl \ Warehouse Topics:}}
                \begin{itemize}[leftmargin=3mm, itemsep=1pt] 
                    \item \tiny Topic 1: Logistics of managing a distribution network  \textit{(Sel.)}
                    \item \tiny Topic 6: Fire Safety Protocols
                    \item \tiny Topic 10: Controlling Inventory Shrinkage
                    \enspace {\tiny \textit{... (+7)}}
                \end{itemize}
            \end{tcolorbox}
            \vspace{1mm}
            {\tiny \textbf{\faQuoteLeft \ Sample:}} \textit{\tiny Context: Logistics of managing a distribution network}
            \par\vspace{1mm}
            \begin{tcolorbox}[enhanced, width=0.75\linewidth, colback=chatuser, frame hidden, arc=1mm, left=1.5mm, top=0.5mm, bottom=0.5mm]
                \tiny \textbf{\color{blue!60!black}\faUserTie\ Importer:} Our sales in Upper Egypt are growing, but delivery from Cairo is slow.
            \end{tcolorbox}
            \vspace{0.5mm}
            \begin{flushright}
            \begin{tcolorbox}[enhanced, width=0.75\linewidth, colback=chatagent, frame hidden, arc=1mm, left=1.5mm, top=0.5mm, bottom=0.5mm]
                \tiny \textbf{\color{blue!60!black}\faUserTie\ Partner:} Agreed. We should set up hubs in Tanta and Assiut.
            \end{tcolorbox}
            \end{flushright}
        \end{tcolorbox}
    \end{minipage}%
    \hfill
    \begin{minipage}[t]{0.49\linewidth}
        \begin{tcolorbox}[datacard={moroccoprimary}{Morocco: Tourism \& Hospitality}{\faPlane}]
            \linespread{0.85}\selectfont
            \vspace{1mm}
            \begin{center}
                \activebadge{moroccoprimary}{moroccohighlight}{\faStore \ Souk}
                \passivebadge{\faSuitcase \ Travel}
                \passivebadge{\faSun \ Safari}
                \enspace {\tiny \textit{... (+52)}}
            \end{center}
            \vspace{1mm}
            \begin{tcolorbox}[enhanced, colback=moroccobg, colframe=moroccobg, borderline west={2pt}{0pt}{moroccohighlight}, arc=1mm, left=1.5mm, top=0.5mm, bottom=0.5mm]
                {\tiny \textbf{\color{moroccoprimary} \faListUl \ Souk Topics:}}
                \begin{itemize}[leftmargin=3mm, itemsep=1pt] 
                    \item \tiny Topic 1: Giving directions to a part of the souk \textit{(Sel.)}
                    \item \tiny Topic 3: Making a counter-offer during haggling
                    \item \tiny Topic 7: Asking for directions to Jemaa el-Fna square
                    \enspace {\tiny \textit{... (+7)}}
                \end{itemize}
            \end{tcolorbox}
            \vspace{1mm}
            {\tiny \textbf{\faQuoteLeft \ Sample:}} \textit{\tiny Context: Giving directions to a part of the souk}
            \par\vspace{1mm}
            \begin{tcolorbox}[enhanced, width=0.75\linewidth, colback=chatuser, frame hidden, arc=1mm, left=1.5mm, top=0.5mm, bottom=0.5mm]
                \tiny \textbf{\color{magenta!80!black}\faFemale\ Tourist:} Excuse me, I'm looking for the spice square. Am I going the right way?
            \end{tcolorbox}
            \vspace{0.5mm}
            \begin{flushright}
            \begin{tcolorbox}[enhanced, width=0.75\linewidth, colback=chatagent, frame hidden, arc=1mm, left=1.5mm, top=0.5mm, bottom=0.5mm]
                \tiny \textbf{\color{blue!60!black}\faUserTie\ Vendor:} You're close. Go straight, then take the second left. You'll find it there.
            \end{tcolorbox}
            \end{flushright}
        \end{tcolorbox}
    \end{minipage}

    \vspace{3mm} 
    
    \noindent
    \begin{minipage}[t]{0.49\linewidth}
        \begin{tcolorbox}[datacard={mauritaniaprimary}{Mauritania: Everyday Social Life}{\faUsers}]
            \linespread{0.85}\selectfont
            \vspace{1mm}
            \begin{center}
                \activebadge{mauritaniaprimary}{mauritaniahighlight}{\faMugHot \ Tea ('Atay')}
                \passivebadge{\faHandshake \ Greet}
                \passivebadge{\faHome \ Family}
                \enspace {\tiny \textit{... (+52)}}
            \end{center}
            \vspace{1mm}
            \begin{tcolorbox}[enhanced, colback=mauritaniabg, colframe=mauritaniabg, borderline west={2pt}{0pt}{mauritaniahighlight}, arc=1mm, left=1.5mm, top=0.5mm, bottom=0.5mm]
                {\tiny \textbf{\color{mauritaniaprimary} \faListUl \ Tea ('Atay') Topics:}}
                \begin{itemize}[leftmargin=3mm, itemsep=1pt] 
                    \item \tiny Topic 1: Deciding whose turn it is to host \& prepare  tea \textit{(Sel.)}
                    \item \tiny Topic 7: The 3 Ceremonial Rounds 
                    \item \tiny Topic 9: Green Tea Quality
                    \enspace {\tiny \textit{... (+7)}}
                \end{itemize}
            \end{tcolorbox}
            \vspace{1mm}
            {\tiny \textbf{\faQuoteLeft \ Sample:}} \textit{\tiny Context: Deciding whose turn it is to host \& prepare  tea}
            \par\vspace{1mm}
            \begin{tcolorbox}[enhanced, width=0.75\linewidth, colback=chatuser, frame hidden, arc=1mm, left=1.5mm, top=0.5mm, bottom=0.5mm]
                \tiny \textbf{\color{magenta!80!black}\faFemale\ Friend 1:} Are you free for tea later? I'm pretty sure it's your turn to host.
            \end{tcolorbox}
            \vspace{0.5mm}
            \begin{flushright}
            \begin{tcolorbox}[enhanced, width=0.75\linewidth, colback=chatagent, frame hidden, arc=1mm, left=1.5mm, top=0.5mm, bottom=0.5mm]
                \tiny \textbf{\color{magenta!80!black}\faFemale\ Friend 2:} No, we were at my place last week. It is definitely your house this time.
            \end{tcolorbox}
            \end{flushright}
        \end{tcolorbox}
    \end{minipage}%
    \hfill
    \begin{minipage}[t]{0.49\linewidth}
        \begin{tcolorbox}[datacard={omanprimary}{Oman: Agriculture \& Farming}{\faLeaf}]
            \linespread{0.85}\selectfont
            \vspace{1mm}
            \begin{center}
                \activebadge{omanprimary}{omanhighlight}{\faWater \ Well Drilling}
                \passivebadge{\faUsers \ Co-ops}
                \passivebadge{\faBug \ Pests}
                \enspace {\tiny \textit{... (+52)}}
            \end{center}
            \vspace{1mm}
            \begin{tcolorbox}[enhanced, colback=omanbg, colframe=omanbg, borderline west={2pt}{0pt}{omanhighlight}, arc=1mm, left=1.5mm, top=0.5mm, bottom=0.5mm]
                {\tiny \textbf{\color{omanprimary} \faListUl \ Well Drilling Topics:}}
                \begin{itemize}[leftmargin=3mm, itemsep=1pt] 
                    \item \tiny Topic 3: Asking for a Guarantee \textit{(Sel.)}
                    \item \tiny Topic 5: Debating which village has the oldest or longest 'falaj'
                    \item \tiny Topic 8: High cost of drilling a private well
                    \enspace {\tiny \textit{... (+7)}}
                \end{itemize}
            \end{tcolorbox}
            \vspace{1mm}
            {\tiny \textbf{\faQuoteLeft \ Sample:}} \textit{\tiny Context: Asking for a Guarantee}
            \par\vspace{1mm}
            \begin{tcolorbox}[enhanced, width=0.75\linewidth, colback=chatuser, frame hidden, arc=1mm, left=1.5mm, top=0.5mm, bottom=0.5mm]
                \tiny \textbf{\color{blue!60!black}\faUserTie\ Farmer:} What happens if we drill and find no water?
            \end{tcolorbox}
            \vspace{0.5mm}
            \begin{flushright}
            \begin{tcolorbox}[enhanced, width=0.75\linewidth, colback=chatagent, frame hidden, arc=1mm, left=1.5mm, top=0.5mm, bottom=0.5mm]
                \tiny \textbf{\color{blue!60!black}\faUserTie\ Driller:} No guarantee. But you pay half if dry.
            \end{tcolorbox}
            \end{flushright}
        \end{tcolorbox}
    \end{minipage}

    \vspace{3mm} 

    \noindent
    \begin{minipage}[t]{0.49\linewidth}
        \begin{tcolorbox}[datacard={palestineprimary}{Palestine: Group Project Work}{\faGraduationCap}]
            \linespread{0.85}\selectfont
            \vspace{1mm}
            \begin{center}
                \activebadge{palestineprimary}{palestinehighlight}{\faClipboardList \ Tawjihi Prep}
                \passivebadge{\faLandmark \ Heritage}
                \passivebadge{\faBookOpen \ Curriculum}
                \enspace {\tiny \textit{... (+52)}}
            \end{center}
            \vspace{1mm}
            \begin{tcolorbox}[enhanced, colback=palestinebg, colframe=palestinebg, borderline west={2pt}{0pt}{palestinehighlight}, arc=1mm, left=1.5mm, top=0.5mm, bottom=0.5mm]
                {\tiny \textbf{\color{palestinehighlight} \faListUl \ Project Work Topics:}}
                \begin{itemize}[leftmargin=3mm, itemsep=1pt] 
                    \item \tiny Topic 1: Tawjihi preparation program \textit{(Sel.)}
                    \item \tiny Topic 2: Palestinian cultural heritage
                    \item \tiny Topic 3: Plan the curriculum
                     \enspace {\tiny \textit{... (+7)}}
                \end{itemize}
            \end{tcolorbox}
            \vspace{1mm}
            {\tiny \textbf{\faQuoteLeft \ Sample:}} \textit{\tiny Context: Tawjihi Preparation Program}
            \par\vspace{1mm}
            \begin{tcolorbox}[enhanced, width=0.75\linewidth, colback=chatuser, frame hidden, arc=1mm, left=1.5mm, top=0.5mm, bottom=0.5mm]
                \tiny \textbf{\color{magenta!80!black}\faFemale\ Admin:} We need to evaluate our Tawjihi Prep program.
            \end{tcolorbox}
            \vspace{0.5mm}
            \begin{flushright}
            \begin{tcolorbox}[enhanced, width=0.75\linewidth, colback=chatagent, frame hidden, arc=1mm, left=1.5mm, top=0.5mm, bottom=0.5mm]
                \tiny \textbf{\color{blue!60!black}\faUserTie\ Head:} Based on results, students struggle with literary subjects.
            \end{tcolorbox}
            \end{flushright}
        \end{tcolorbox}
    \end{minipage}%
    \hfill
    \begin{minipage}[t]{0.49\linewidth}
        \begin{tcolorbox}[datacard={saudiprimary}{Saudi: Construction \& Real Estate}{\faHardHat}]
            \linespread{0.85}\selectfont
            \vspace{1mm}
            \begin{center}
                \activebadge{saudiprimary}{saudihighlight}{\faDraftingCompass \ Architectural Eng.}
                \passivebadge{\faHardHat \ Site Ops}
                \passivebadge{\faCity \ Urban}
                \enspace {\tiny \textit{... (+52)}}
            \end{center}
            \vspace{1mm}
            \begin{tcolorbox}[enhanced, colback=saudibg, colframe=saudibg, borderline west={2pt}{0pt}{saudiprimary}, arc=1mm, left=1.5mm, top=0.5mm, bottom=0.5mm]
                {\tiny \textbf{\color{saudiprimary} \faListUl \ Architectural Topics:}}
                \begin{itemize}[leftmargin=3mm, itemsep=1pt] 
                    \item \tiny Topic 2: Reviewing structural integrity of formwork \textit{(Sel.)}
                    \item \tiny Topic 1: Design a home with a separate entrance
                    \item \tiny Topic 3: Inclusion of home cinema in basement design
                    \enspace {\tiny \textit{... (+7)}}
                \end{itemize}
            \end{tcolorbox}
            \vspace{1mm}
            {\tiny \textbf{\faQuoteLeft \ Sample:}} \textit{\tiny Context: Reviewing structural integrity of formwork}
            \par\vspace{1mm}
            \begin{tcolorbox}[enhanced, width=0.75\linewidth, colback=chatuser, frame hidden, arc=1mm, left=1.5mm, top=0.5mm, bottom=0.5mm]
                \tiny \textbf{\color{blue!60!black}\faUserTie\ Eng:} The supports in the central bay seem under-spaced. Check the drawings.
            \end{tcolorbox}
            \vspace{0.5mm}
            \begin{flushright}
            \begin{tcolorbox}[enhanced, width=0.75\linewidth, colback=chatagent, frame hidden, arc=1mm, left=1.5mm, top=0.5mm, bottom=0.5mm]
                \tiny \textbf{\color{blue!60!black}\faUserTie\ Foreman:} Yes, Engineer. I see the note. We'll add extra props immediately.
            \end{tcolorbox}
            \end{flushright}
        \end{tcolorbox}
    \end{minipage}

    \caption{Examples of the topic and conversation generation process across six countries.  The process defines 55 high-level subdomains, each expanding into 10 specific topics along with their personas (gender \& roles); a conversation is then generated for each topic-persona pair.}
   
    \label{fig:six_country_cards}
\end{figure*}
\newcommand{\arb}[1]{\RL{#1}}
\newcommand{\err}[1]{{\color{red}\RL{#1}}}
\definecolor{fixx}{rgb}{0.1, 0.5, 0.1}
\newcommand{\ARcell}[1]{%
  \begin{tabular}[t]{@{}r@{}}#1\end{tabular}%
}

\begin{table*}[t!]
\centering
\footnotesize
\renewcommand{\arraystretch}{1.6}
\setlength{\tabcolsep}{5pt}


\begin{tabularx}{\textwidth}{
>{\raggedright\arraybackslash}p{2.0cm}
>{\raggedright\arraybackslash}p{3.5cm}
>{\raggedright\arraybackslash}X
>{\raggedright\arraybackslash}X
c}
\toprule
\textbf{Country} &
\textbf{English Source Sentence} &
\textbf{Pre-revised Translation} &
\textbf{Revised Translation} &
\textbf{Speaker} \\
\midrule


Yemen &
Please look up at the half-moon windows above the doorway. &
\AR{لو <\textcolor{red}{\<سمحتم>}> شوفوا للطاقات اللي على شكل هلال فوق المدخل.} & 
\AR{لو <\textcolor{fixx}{\<سمحتي>}> شوفي للطاقات اللي على شكل هلال فوق البوابة.} &
{\color{magenta}\faFemale}$\to${\color{magenta}\faFemale} \\
\midrule

Saudi Arabia &
Here is the problem. You can see the water dripping from this pipe behind the washing machine. &
\AR{هنا المشكلة.  <\textcolor{red}{\<تشوف>}>  <\textcolor{red}{\< الماي ينقط>}> من <\textcolor{red}{\<هالهوز>}> <\textcolor{red}{\<ورا>}>الي  <\textcolor{red}{\<الغسالة>}>.} &
\AR{هنا المشكلة، <\textcolor{fixx}{\<شايف 
  الموية تنقط>}> من  <\textcolor{fixx}{\<هذي الماصورة >}> <\textcolor{fixx}{\<ورى>}> <\textcolor{fixx}{\<المغسلة>}>. }  &
{\color{blue}\faMale}$\to${\color{blue}\faMale} \\
\midrule

Palestine &
Thank you very much. I’m glad I could contribute, especially with coordinating the user feedback sessions with the development team. &
\AR{شكراً كثير الك. مبسوطة إني قدرت أساهم، وخصوصاً في تنسيق جلسات <\textcolor{red}{feedback}> <\textcolor{red}{user}> مع فريق <\textcolor{red}{development}>.} &
\AR{شكراً كثير الك. مبسوطة إني قدرت أساهم، وخصوصاً في تنسيق   <\textcolor{fixx}{\<جلسات   >}> <\textcolor{fixx}{\<التغذية>}> <\textcolor{fixx}{\<الراجعة>}> للمستخدمين مع فريق <\textcolor{fixx}{\<التطوير>}>.} &
{\color{magenta}\faFemale}$\to${\color{blue}\faMale} \\
\midrule

Jordan &
Welcome aboard! Let’s go over the payments. The company’s commission is 25\% on every trip. &
\AR{أهلاً فيك معنا! خلينا نحكي عن الدفعات. عمولة الشركة <\textcolor{red}{\<\% 25>}> على كل رحلة.} &
\AR{أهلاً فيك معنا! خلينا نحكي عن الدفعات. عمولة الشركة   <\textcolor{fixx}{\<خمسة وعشرين بالمية>}>على كل رحلة.} &
{\color{magenta}\faFemale}$\to${\color{blue}\faMale} \\
\midrule

Oman &
Good morning. That’s a wise precaution. I can come next Tuesday. We will inject each tree directly to protect it. &
\AR{صباح الخير، زين سويتي كذا، اقدر <\textcolor{red}{\<اجيك>}> يوم الثلاثاء الجاي، انضرب كل نخلة بمرة.} &
\AR{صباح الخير، زين سويتي كذا، اقدر <\textcolor{fixx}{\<اجيش>}> يوم الثلاثاء الجاي، <\textcolor{fixx}{\<عشان نحميها>}>انضرب كل نخلة بمرة .} &
{\color{blue}\faMale}$\to${\color{magenta}\faFemale} \\
\bottomrule
\end{tabularx}

\caption{Illustrative examples of prerevised and revised Arabic translations from the human-only revision phase across different countries.}
\label{tab:revision_comparison}
\end{table*}
\begin{table*}[t!]
\centering
\footnotesize 
\renewcommand{\arraystretch}{1.6} 
\setlength{\tabcolsep}{5pt}

\begin{tabularx}{\textwidth}{>{\centering\arraybackslash}p{2.3cm} X >{\raggedright\arraybackslash}X c}
\toprule
\rowcolor{gray!15}
\textbf{Dialect / Domain} & \textbf{English Context (Source)} & \textbf{Dialectal Arabic (Target)} & \textbf{Speaker} \\ \midrule

\multirow{2}{*}{\shortstack[c]{\activebadge{mauritaniaprimary}{mauritaniaprimary}{Sudan} \\ \tiny {\color{violet}\faUsers} \\ \tiny Everyday \& social life}} 
& I've started making a big jug of cold hibiscus tea every morning. It's the only way to get through this heat. & \AR{انا بديت أعمل جك عصير كركدي كبير كل صباح، دي الطريقة الوحيدة مع السخانة دي} & {\color{magenta}\faFemale}$\to${\color{blue}\faMale} \\ \cdashline{2-4}
& That's a very good idea. I should do that for the children. They get so thirsty. & \AR{دي فكرة سمحة شديد، لازم عشان كده للأولاد، بيعطشو شديد} & {\color{blue}\faMale}$\to${\color{magenta}\faFemale} \\ \midrule

\multirow{3}{*}{\shortstack[c]{\activebadge{egyptprimary}{egyptprimary}{Egypt} \\ \tiny {\color{brown}\faTools} \\ \tiny Construction \& real estate}} 
& Ahmed. The container just cleared customs in Alexandria. Can you have a truck ready to load it tomorrow morning? & \AR{ يا احمد، الصندوق عدى حالا من الجمارك في اسكندرية. تقدر تجهز مقطورة بكرة الصبح عشان نحمله؟} & {\color{blue}\faMale}$\to${\color{blue}\faMale} \\ \cdashline{2-4}
& Yes, of course. I have a truck available. Send me the release order, and we'll be there first thing, God willing. & \AR{ايوة طبعا. انا عندي مقطورة. ابعتلي رقم امر الافراج وهنبقى هناك من النجمة بأمر الله.} & {\color{blue}\faMale}$\to${\color{blue}\faMale} \\ \midrule

\multirow{3}{*}{\shortstack[c]{\activebadge{moroccoprimary}{moroccoprimary}{Morocco} \\ \tiny {\color{cyan}\faHeadset} \\ \tiny Professional \& workplace}} 
& Hello, IT support, how may I help you? & \AR{الو، <Assistance informatique>، كيفاش نقدر نعاونك؟} & {\color{blue}\faMale}$\to${\color{magenta}\faFemale} \\ \cdashline{2-4}
& Hello, I can't log into my account. I think I've forgotten my password. & \AR{الو، منقدرش ندخل لل <compte> ديالي. واقيلا نسيت ل <mot de passe> ديالي.} & {\color{magenta}\faFemale}$\to${\color{blue}\faMale} \\ \cdashline{2-4}
& No problem at all, we can sort this out. Are you in front of your computer right now? & \AR{ماشي مشكل غنحاولو نلقاو الحل. واش انتي حدا ل <pc> دابا؟} & {\color{blue}\faMale}$\to${\color{magenta}\faFemale} \\ \midrule

\multirow{2}{*}{\shortstack[c]{\activebadge{mauritaniaprimary}{mauritaniaprimary}{Mauritania} \\ \tiny {\color{orange}\faShoppingCart} \\ \tiny Commerce \& transactions}} 
& Excuse me, is the hibiscus juice made fresh here? & \AR{عفوا، يعدل هون عصير امبصام اجديد؟} & {\color{magenta}\faFemale}$\to${\color{blue}\faMale} \\ \cdashline{2-4}
& Yes, madam. We prepare it fresh every morning. & \AR{أهيه، <madam>. نحن انعدلوه كل صبحاية اجديد.} & {\color{blue}\faMale}$\to${\color{magenta}\faFemale} \\ \midrule

\multirow{2}{*}{\shortstack[c]{\activebadge{moroccoprimary}{moroccoprimary}{Lebanon} \\ \tiny {\color{darkgray}\faSeedling} \\ \tiny Agriculture \& farming}} 
& Here, I saved some of my local zucchini seeds from last year's harvest. They have the best flavor. & \AR{هون، انا احتفظت بشوي بزر كوسى البلدي تعولي من موسم السنة الماضية. طعمتا أطيب شي.} & {\color{magenta}\faFemale}$\to${\color{magenta}\faFemale} \\ \cdashline{2-4}
& Oh, thank you so much! You're a lifesaver. These are much better than the ones they sell at the store. & \AR{يا سلام، شكرا كتير! أنقذتني. هودي أحسن بكتير من يلي عم يبيعوهن بالسوق.} & {\color{magenta}\faFemale}$\to${\color{magenta}\faFemale} \\ 

\bottomrule
\end{tabularx}
\caption{Cross-dialectal dialogue examples from Alexandria dataset.}
\label{tab:alexandria_qualitative_final}
\end{table*}
\begin{table}[!ht]
    \centering
    \small
    \setlength\dashlinedash{0.5pt} 
    \setlength\dashlinegap{1.5pt}  
    \begin{tabularx}{\columnwidth}{@{}l X@{}}
        \toprule
        \textbf{Country} & \textbf{Covered Subdialects} \\
        \midrule
        Egypt & Egyptian Arabic (Cairene) \\
        \hdashline
        Jordan & Jordanian Irbidi \\
        \hdashline
        Lebanon & Lebanese Standard \\
        \hdashline
        Libya & Libyan Arabic (Misrati/Central) \\
        \hdashline
        Mauritania & Mauritanian Hassaniya \\
        \hdashline
        Morocco & Moroccan Standard Darija Dialect \\
        \hdashline
        Oman & Omani Al-Wafi \\
             & Omani Ibri (Al Nahda) \\
             & Omani Rustaqi \\
             & Omani Seebi (Al Mawaleh) \\
             & Omani Suri (Bani Khuzaymah) \\
        \hdashline
        Palestine & Palestinian Albira (Urban) \\
                  & Palestinian Arabic (Aboud Falahi) \\
                  & Palestinian Arabic (Kobar Falahi) \\
                  & Palestinian Arabic (Ni'lin Falahi) \\
                  & Palestinian Arabic (Noba Falahi) \\
                  & Palestinian Arabic (Ramallah Falahi) \\
                  & Palestinian Arabic (Shuqba Falahi) \\
                  & Palestinian Arabic (Silwad Falahi) \\
                  & Palestinian Arabic (Surif Falahi) \\
                  & Palestinian Nabulsi (Urban) \\
        \hdashline
        Saudi Arabia & Saudi Arabic (Southern) \\
                      & Saudi Arabic Hijazi \\
                      & Saudi Arabic Khaleeji \\
        \hdashline
        Sudan & Sudanese Standard \\
        \hdashline
        Syria & Syrian Arabic (Homsi) \\
              & Syrian Arabic (Levantine Standard) \\
        \hdashline
        Tunisia & Tunisian \\
        \hdashline
        Yemen & Yemeni Arabic (Central) \\
              & Yemeni San'ani \\
              & Yemeni Taiz \\
        \bottomrule
    \end{tabularx}
    \caption{Arabic Subdialects by Country covered in the Alexandria project.}
    \label{tab:arabic_dialects}
\end{table}

\subsection{Annotation Platform}

The entire data collection and revision workflow was executed using a spreadsheet-based infrastructure (Google Sheets). The generated English conversations were shuffled and partitioned into batches of 300 conversations (approximately 1,000 turns), with each batch exported to a dedicated sheet. Once a participant finishes a sheet, we can assign them another sheet.

\begin{figure*}[!ht]
    \centering
    \includegraphics[width=\textwidth]{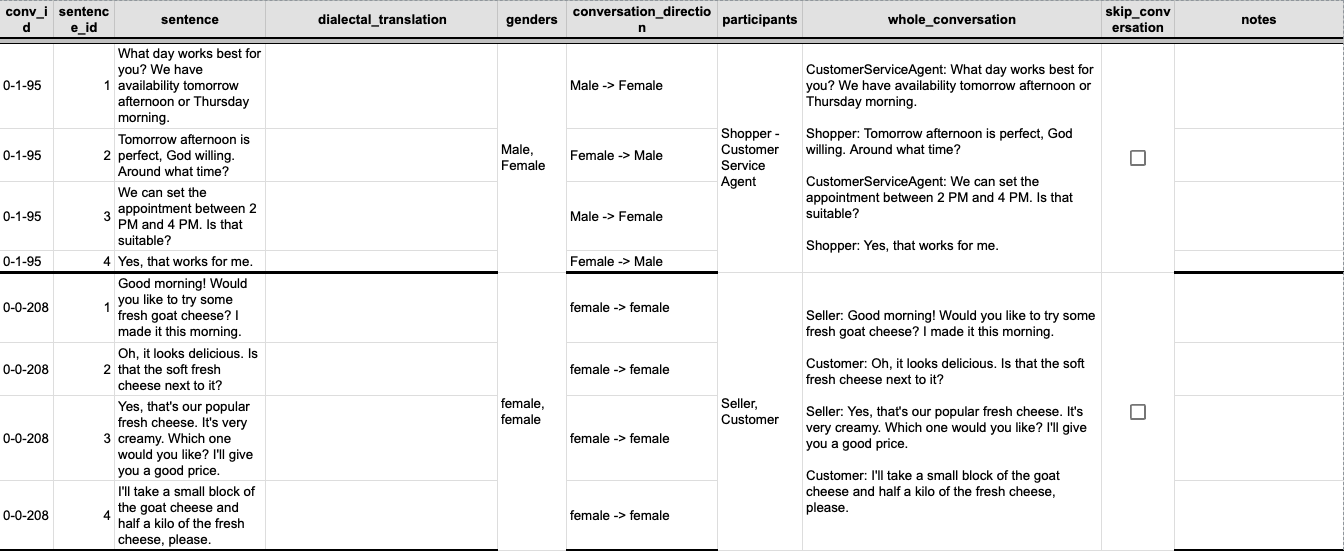}
    \caption{Screenshot of the Translation Interface (Google Sheets). Translators are provided with the English source, gender direction, and persona metadata, and enter the dialectal translation in the designated column.}
    \label{fig:translation_sheet_ui}
\end{figure*}

\paragraph{Translation Interface.} Each conversation was annotated with metadata indicating the participating personas and the gender direction for each turn. The interface provided translators with a checkbox to discard an entire conversation if any constituent sentence was deemed irrelevant or problematic. Translators entered their translations in a designated column, strictly adhering to the specified gender and social register. An additional field was provided for translators to log specific notes or linguistic observations for each turn. Figure \ref{fig:translation_sheet_ui} shows a screenshot from one of the translation sheets.

\begin{figure*}[!ht]
    \centering
    \includegraphics[width=\textwidth]{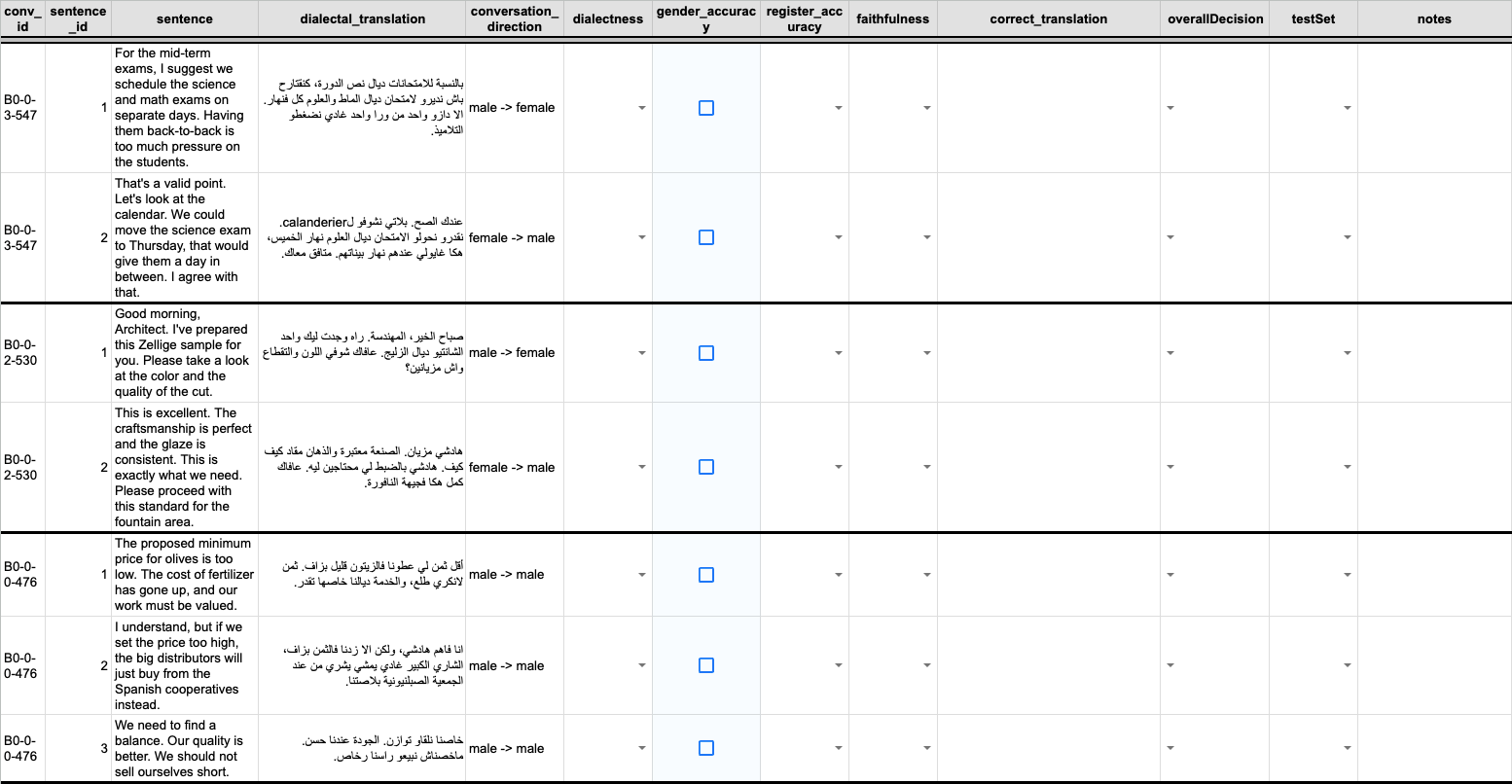}
    \caption{Screenshot of the Peer-Revision Interface. Reviewers assess translations based on dialectness, gender accuracy, register, and faithfulness, and provide corrections where necessary.}
    \label{fig:revision_sheet_ui}
\end{figure*}

\paragraph{Revision Interface.} For the peer-revision phase, reviewers received separate sheets containing the source English text and the anonymized dialectal translations collected during the previous phase. Reviewers were tasked with populating specific columns for quality scores and, where necessary, providing corrected translations. A notes column was also available for qualitative feedback. Figure \ref{fig:revision_sheet_ui} shows a screenshot from one of the revision sheets.

\paragraph{Access Control and Monitoring.} To ensure data integrity and privacy, access to each sheet was strictly limited to the assigned participant and their team leader. We implemented a centralized progress tracking system (on Google Sheets) that aggregated statistics from all individual sheets. This system logged daily metrics to monitor throughput at both the country and participant levels. Additionally, team leaders were provided with a customized dashboard view of this global report, enabling real-time oversight of their respective teams' progress.

\subsection{AI Usage in Translation Phase}~\label{app:ai_use_translation}

\paragraph{AI as an Auxiliary Assistance Tool} 
To preserve the authenticity of the dialectal data, our protocol strictly defined generative AI as an \textit{auxiliary assistance tool} rather than a primary translation source. The initial protocol prioritized fully manual translation; however, we refined this policy to allow a machine-translation-assisted workflow specifically for technical domains such as \textbf{Energy, Mining, and Logistics}. In these cases, AI served as a comprehension and lexical aid to help translators "bootstrap" initial MSA drafts. This workflow was permitted only under the condition that the final output was a product of rigorous manual post-editing and adaptation to ensure dialectal integrity. 

\paragraph{Adoption and Frequency of Assistance} 
Based on a survey conducted with \textbf{28 participants}, the adoption of these auxiliary tools was widespread; self-reported data indicates that 85.7\% of participants utilized an AI system or translation utility as part of their workflow, while only 14.3\% relied solely on fully manual translation. Among those who utilized AI for assistance, the degree of reliance varied. While 29.2\% of users reported low reliance (1--100 sentences) and an equal percentage reported moderate reliance (100--500 sentences), 37.5\% indicated high reliance exceeding 500 sentences for initial drafting.

\paragraph{Tool Selection for Assistance} 
Participants often employed multiple systems simultaneously to verify assistance outputs. Google Translate was the dominant tool, utilized by 87.5\% of AI users (21 participants), primarily for its speed in retrieving technical terminology. LLMs served as frequent supplementary aids, with various versions of ChatGPT used by 41.7\% of the cohort. Other tools used strictly for lexical retrieval included online dictionaries like Cambridge or Linguee, and alternative LLMs such as Gemini or Qwen.

\paragraph{Primary Modes of Assistance} 
The specific modes of assistance identified by participants reinforce that AI served as a lexical and comprehension bridge rather than a replacement for human translation. The most frequent applications included lexical support for specific technical terms (21 mentions) and improving the comprehension of complex source English text (15 mentions). AI was used significantly less for drafting entire sentences (7 mentions), and in all such cases, these drafts served as "base" versions for human-led dialectal adaptation.

\paragraph{The MSA Bridge Strategy} 
A critical finding from translator feedback was the emergence of the \textit{MSA Bridge Strategy}, where AI functioned as an intermediate step. Because commercial AI models frequently struggle to produce authentic local dialects, translators used AI to generate an intermediate MSA version of the text to ensure technical accuracy. This strategy ensured that while AI provided the technical "bridge," the linguistic integrity and final dialectal variety remained entirely human-validated.

\subsection{Qualitative Analysis of Translation Challenges} \label{app:translation_challenges}

We conducted a qualitative survey of the translators to identify linguistic and non-linguistic friction points in the English-to-Dialectal Arabic translation pipeline. The feedback highlights three primary categories of challenges:

\paragraph{Lexical Gaps and Domain Specificity} A significant hurdle reported by annotators was the lack of direct dialectal equivalents for technical and specialized terminology (e.g., in mining, geology, or corporate logistics). Dialectal Arabic is predominantly a spoken register used for daily communication; consequently, translators frequently resorted to MSA or code-switching to convey scientific concepts. Where direct equivalents were absent, translators utilized periphrasis, replacing single English words with descriptive phrases, which introduced structural divergence between the source and target.

\paragraph{Fidelity vs. Fluency Trade-offs} The annotation guidelines' requirement for strict semantic faithfulness often conflicted with the goal of producing natural, conversational dialect. Annotators noted that preserving the syntactic structure of English resulted in "translationese"—phrasing that is grammatically correct but pragmatically unnatural in a dialectal context. Idioms and fixed expressions proved particularly difficult to map, requiring significant rewording to maintain the original intent without sacrificing the colloquial tone.

\paragraph{Source Ambiguity and Sociocultural Mismatch} Issues inherent to the source text further complicated the process. Annotators cited ambiguity and vague references in the English source as a cause for interpretation delays. Furthermore, cultural disparities posed a distinct challenge; scenarios depicting gender roles uncommon in the target culture (e.g., female electricians or delivery personnel) were perceived as unnatural to the Arab world. This highlights a critical sociolinguistic challenge: accurately translating the \textit{meaning} of a sentence while navigating the \textit{cultural expectations} embedded in the target dialect.

\section{Evaluation}\label{app:eval_setup}

\begin{table}[ht!]
\centering
\resizebox{\columnwidth}{!}{%
    \small
    \renewcommand{\arraystretch}{1.25}
    \begin{tabular}{llr}
    \toprule
    \textbf{Category} & \textbf{Model} & \textbf{Size} \\
    \midrule
    \multicolumn{3}{l}{\textbf{Closed-source LLMs}} \\
    \midrule
  
    Google  & Gemini-3-Pro     & N/A \\
            & Gemini-3-Flash & N/A \\
            & Gemini-2.5-Pro & N/A \\
            & Gemini-2.5-Flash & N/A \\

    \midrule
    \multicolumn{3}{l}{\textbf{Open-source LLMs}} \\
    \midrule
    Google  & Gemma-3-27B-IT~\cite{gemma_2025}     & 27B \\
            & Gemma-3-12B-IT~\cite{gemma_2025}    & 12B \\
            & Gemma-3-4B-IT~\cite{gemma_2025}    & 4B \\
    \hdashline
    Cohere  & Command-A~\cite{cohere2025commandaenterprisereadylarge} & 111B \\
            & Command-A-Translate~\cite{kocmi-etal-2025-command} & 111B \\
            & Command-A-Reasoning~\cite{cohere2025commandaenterprisereadylarge} & 111B \\
            & Command-R7B-Arabic~\cite{alnumay2025command}  & 7B \\
            & Aya-Expanse-32B~\cite{dang2024ayaexpansecombiningresearch}      & 32B \\
            & Aya-Expanse-8B~\cite{dang2024ayaexpansecombiningresearch}       & 8B \\
    \hdashline
    Humain  & ALLAM-7B-Instruct~\cite{bari2025allam}       & 7B \\
    \hdashline
    QCRI    & Fanar-1-9B-Instruct~\cite{fanarteam2025fanararabiccentricmultimodalgenerative}    & 9B \\
    \hdashline
    Qwen    & Qwen3-Next-80B-A3B~\cite{yang2025qwen3technicalreport}  & 80B \\
            & Qwen3-32B~\cite{yang2025qwen3technicalreport}           & 32B \\
            & Qwen3-8B~\cite{yang2025qwen3technicalreport}            & 8B \\
            & Qwen3-4B~\cite{yang2025qwen3technicalreport}            & 4B \\
    \hdashline
    OpenAI  & GPT-OSS-120B~\cite{openai2025gptoss120bgptoss20bmodel}        & 120B \\
            & GPT-OSS-20B~\cite{openai2025gptoss120bgptoss20bmodel}         & 20B \\
    \hdashline
    Meta    & Llama-3.3-70B-Instruct~\cite{grattafiori2024llama3herdmodels} & 70B \\
    \bottomrule
    \end{tabular}%
}
\caption{List of Arabic-aware open-source and closed-source LLMs evaluated with Alexandria test set.}
\label{tab:evaluated_llms}
\end{table}

\begin{figure*}[t]
    \centering
    
    \begin{promptbox}[title=Turn-Level Prompt Configuration]
Translate the English text contained in the JSON input into <DIALECT>.

Input:
\{
    "country": "<Country>",
    "domain": "<Domain>",
    "participants": ["<Speaker1>", "<Speaker2>"],
    "gender\_direction": "<Gender>",
    "speaker": "<Current\_Speaker>",
    "text": "<Source\_Text>"
\}

Guidelines:\\
- Return the result strictly in valid JSON.\\
- Translate to <DIALECT> using Arabic script.\\
- Do not add any code, explanations, comments, or any other extra text.\\
- Keep the meaning and tone and respect the gender direction.\\
- Consider the country, the domain, the participants, and the speaker in your translation.

Output scheme:
\{ "translation": "translated text here" \}
    \end{promptbox}
    \vspace{0.2cm}

    \begin{promptbox}[title=Context-Level Prompt Configuration]
Translate the given turn of a conversation from English to <DIALECT>, considering the previous context if provided.

Input:
\{
    "country": "<Country>",
    "domain": "<Domain>",
    "participants": [...],
    "context": ["<Previous\_Turn\_1>", "<Previous\_Turn\_2>"],
    "current\_turn": \{ "speaker": "...", "text": "..." \}
\}

Guidelines:\\
... [Same as Turn-Level] ...\\
- Only translate the "text" field of the "current\_turn".\\
- If a context is provided, do not translate it, and use it to inform your translation.

Output scheme:
\{ "translation": "translation of the text from the current turn" \}
    \end{promptbox}
    \vspace{0.2cm}

    \begin{promptbox}[title=Conversation-Level Prompt Configuration]
Translate all turns in the following conversation from English to <DIALECT>.

Input:
\{
    "country": "<Country>",
    "domain": "<Domain>",
    "participants": [...],
    "turns": [
        \{"speaker": "A", "text": "..."\},
        \{"speaker": "B", "text": "..."\}
    ]
\}

Guidelines:\\
... [Same as Turn-Level] ...

Output scheme:
\{
    "turn\_1": "translation of the text from turn\_1",
    "turn\_2": "translation of the text from turn\_2",
    ...
\}
    \end{promptbox}

    \caption{The three prompt configurations used for the English $\to$ Arabic Dialect evaluation. Note that for the reverse direction (Dialect $\to$ English), the source/target languages are swapped, and the guideline regarding \textbf{Arabic script} is removed.}
    \label{fig:llm_evaluation_prompts}
\end{figure*}

\subsection{Human Evaluation}\label{app:human_eval}

Our human evaluation assessed English-to-Dialect translations across three decoupled dimensions: \textit{Semantic Adequacy}, \textit{Gender Accuracy}, and \textit{Dialectness \& Fluency}. Native speakers of the target dialects followed the specific scoring protocols detailed below.

\subsubsection{Semantic Adequacy (XSTS)}
Annotators evaluated meaning preservation using a 5-point Crosslingual Semantic Textual Similarity (XSTS) scale \cite{agirre-etal-2012-semeval}. They were instructed to ignore grammar, style, or dialect errors for this metric.

\begin{description}
    \item[5 (Perfect)] Meaning is identical; all nuances and tone are preserved.
    \item[4 (Good)] Core meaning is correct; minor nuances (e.g., \textit{huge} vs. \textit{big}) are lost.
    \item[3 (Acceptable)] Main message conveyed; non-critical details missing or slightly inaccurate.
    \item[2 (Poor)] Critical information is missing or wrong; meaning is significantly altered.
    \item[1 (Wrong)] Unrelated to source, contradictory, or gibberish.
\end{description}

\subsubsection{Gender Accuracy}
Annotators verified adherence to the specified grammatical gender direction (e.g., Male speaker $\to$ Female listener).

\begin{description}
    \item[Pass (1)] Correct use of gendered forms (pronouns, verbs, adjectives).
    \item[Fail (0)] Incorrect gender marking (e.g., masculine \textit{anta} instead of feminine \textit{anti}).
    \item[N/A] Sentence is gender-neutral; no specific markers required.
\end{description}

\subsubsection{Dialectness \& Fluency}
Annotators assessed the output by answering the specific question: \textit{``Does this sound like a native speaker of the target dialect (e.g., Moroccan, Levantine)?''}

\begin{description}
    \item[5 (Native)] 100\% Authentic. Uses slang/idioms correctly; contains no MSA.
    \item[4 (Good)] Correct dialect grammar. Phrasing is a bit stiff, but clearly local.
    \item[3 (Hybrid)] Mixes Dialect and MSA. Phrasing feels awkward or ``translated.''
    \item[2 (MSA)] Correct Arabic, but it is Formal (MSA), not Dialect.
    \item[1 (Fail)] Gibberish, wrong dialect entirely, or not Arabic.
\end{description}

\subsubsection{Protocol for MSA Leakage}
To isolate meaning from register control, annotators were instructed to score semantic adequacy and dialectness independently.
\begin{quote}
    \textit{Example:} A correct MSA translation for a request in Moroccan Arabic receives a \textbf{Semantic Score of 5} (perfect meaning) but a \textbf{Dialect Score of 1--2} (wrong register). XSTS scores are not penalized for dialect errors.
\end{quote}

\section{Results}

\subsection{Automatic Evaluation Results} \label{app:automatic_results}

\begin{figure*}[!ht]
    \centering
    \includegraphics[width=\textwidth]{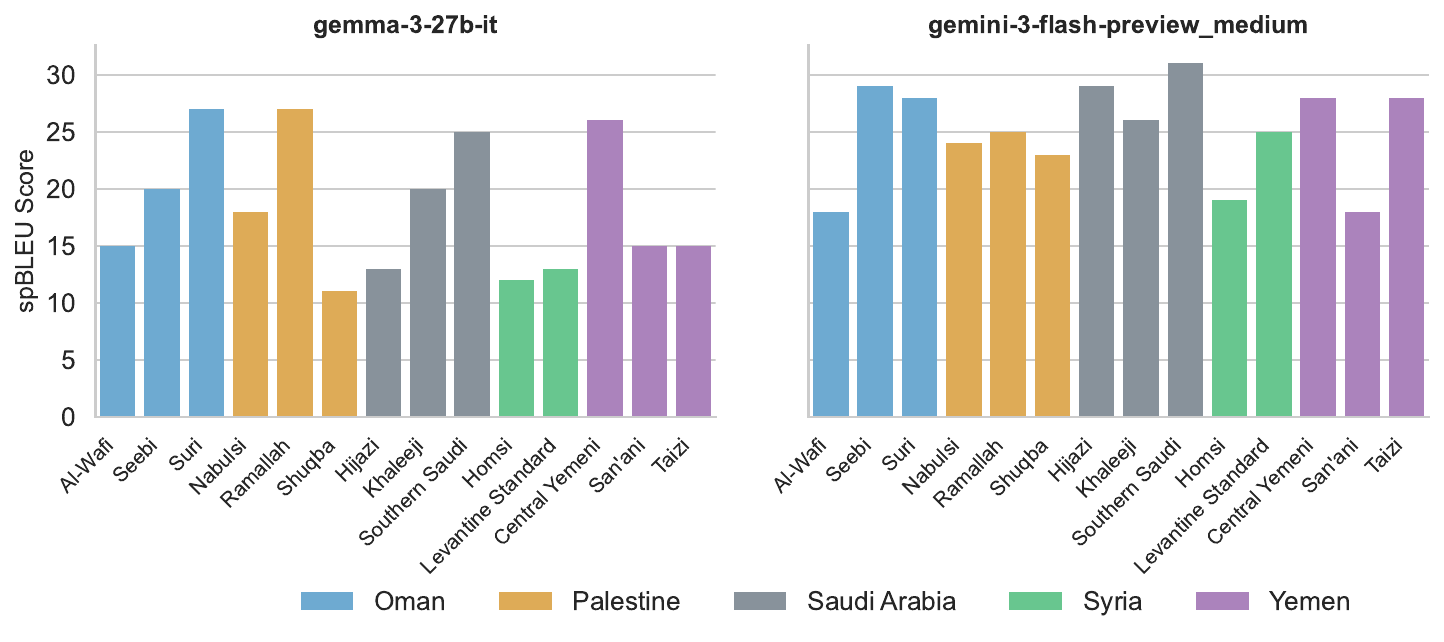}
    \caption{spBLEU scores for selected LLMs on the Alexandria test set (English $\to$ Sub-Dialect). We report results for specific sub-dialects across five countries to highlight intra-country performance discrepancies.}
    \label{fig:subdialects_scores_app}
\end{figure*}

\begin{figure*}[!ht]
    \centering
    \includegraphics[width=\textwidth]{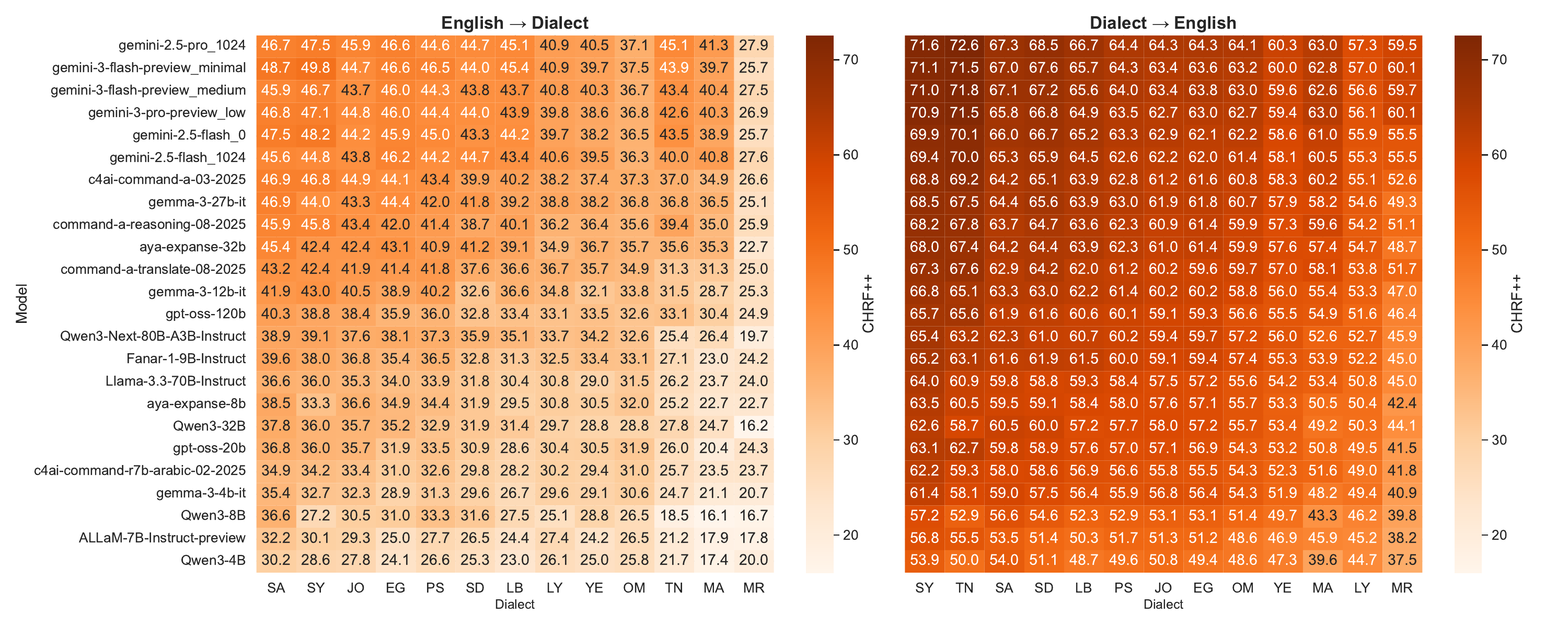}
    \caption{chrF++ scores for LLM-based machine translation on the Alexandria test set. Results cover 13 dialects in both directions (English $\to$ Dialect and Dialect $\to$ English).}
    \label{fig:scores_chrf_heatmap}
\end{figure*}

\begin{figure}[!ht]
    \centering
    \includegraphics[width=0.49\textwidth]{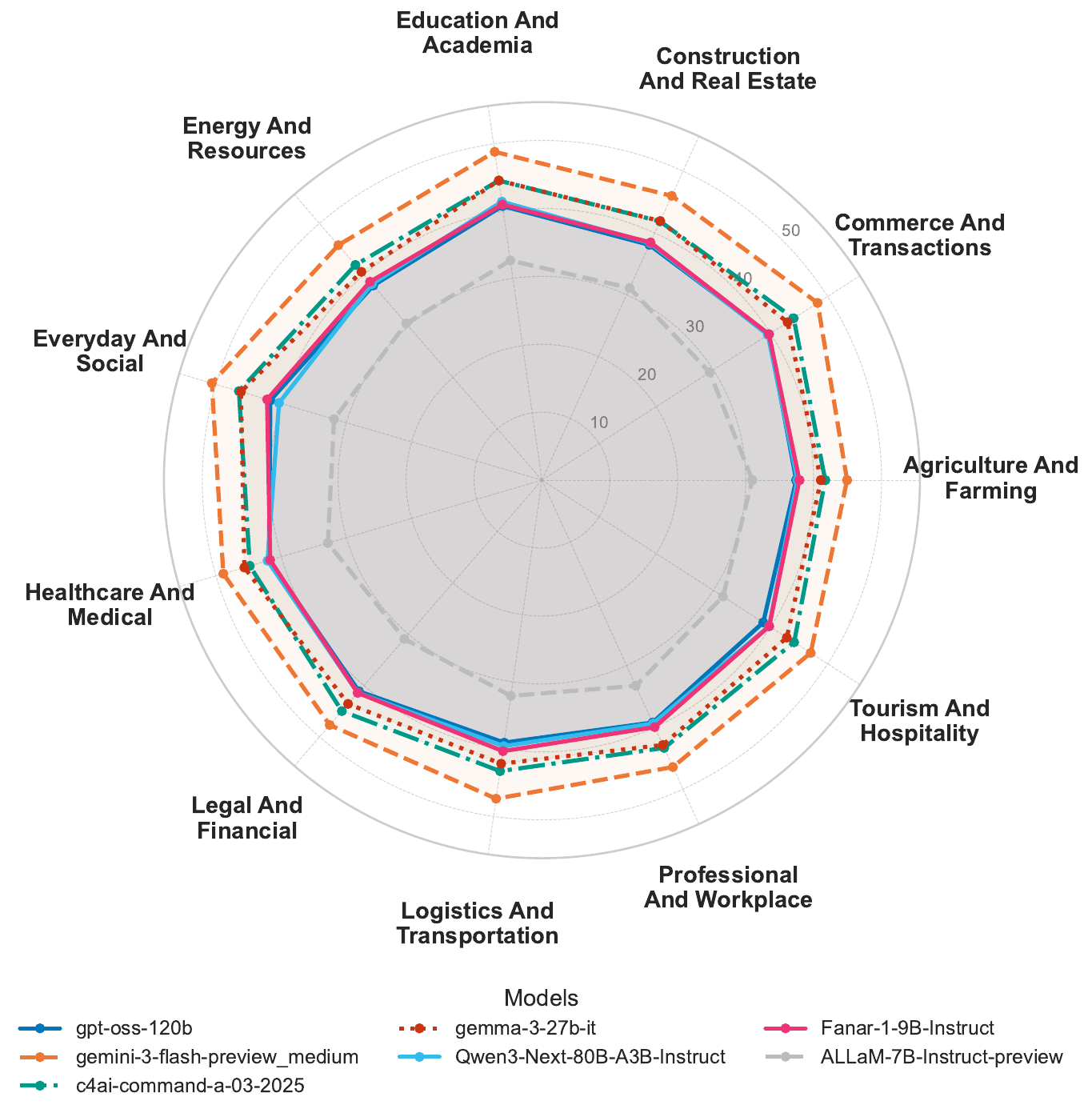}
    \caption{Domain robustness analysis (Dialect $\to$ English). The radar chart illustrates spBLEU scores for a subset of models across all 11 domains, demonstrating consistent performance stratification regardless of the topic.}
    \label{fig:domain_to_en_robust}
\end{figure}

\subsection{Human Evaluation Results}\label{app:human_eval_results}

\begin{table*}[h]
\centering
\resizebox{\textwidth}{!}{%
\begin{tabular}{lccccccccccccc}
\toprule
\textbf{Model} & \textbf{EG} & \textbf{JO} & \textbf{LB} & \textbf{LY} & \textbf{MA} & \textbf{MR} & \textbf{OM} & \textbf{PS} & \textbf{SA} & \textbf{SD} & \textbf{SY} & \textbf{TN} & \textbf{YE} \\ \midrule
gemini-3-flash-preview\_medium & 97.5 & 100.0 & 100.0 & 99.4& 100.0 & 99.4 & 98.9 & 100.0 & 99.6 & 100.0 & 95.2 & 98.8 & 100.0 \\
gemini-3-flash-preview\_minimal & 95.1 & 100.0 & 99.6 & 99.4& 100.0 & 99.4 & 96.1 & 99.6 & 99.6 & 100.0 & 97.6 & 95.1 & 100.0 \\
c4ai-command-a-03-2025 & 96.3 & 97.7 & 98.4 & 100.0 & 100.0 & 99.4 & 98.9 & 100.0 & 99.6 & 98.8 & 95.8 & 95.1 & 96.2 \\
gpt-oss-120b & 93.8 & 100.0 & 98.8 & 96.4& 98.2 & 99.4 & 98.4 & 99.6 & 99.2 & 98.8 & 91.7 & 92.6 & 100.0 \\
gemma-3-27b-it & 91.4 & 97.7 & 96.4 & 97.6& 100.0 & 97.1 & 98.9 & 99.2 & 98.4 & 95.2 & 91.1 & 96.3 & 100.0 \\
aya-expanse-32b & 92.6 & 100.0 & 96.0 & 96.4& 98.8 & 97.1 & 97.7 & 98.8 & 98.4 & 98.8 & 86.3 & 96.3 & 98.8 \\ \bottomrule
\end{tabular}%
}
\caption{Human Evaluation Gender Accuracy (Pass \%) across different countries}
\label{tab:gender_accuracy_results}
\end{table*}
\begin{table*}[h]
\centering
\resizebox{\textwidth}{!}{%
\begin{tabular}{lccccccccccccc}
\toprule
\textbf{Model} & \textbf{EG} & \textbf{JO} & \textbf{LB} & \textbf{LY} & \textbf{MA} & \textbf{MR} & \textbf{OM} & \textbf{PS} & \textbf{SA} & \textbf{SD} & \textbf{SY} & \textbf{TN} & \textbf{YE} \\ \midrule
gemini-3-flash-preview\_medium & 3.94 & 5.00 & 4.78 & 4.03& 4.73 & 3.13 & 4.85 & 4.87 & 4.98 & 4.64 & 4.07 & 3.21 & 4.74 \\
gemini-3-flash-preview\_minimal & 3.86 & 5.00 & 4.78 & 4.01& 4.65 & 3.13 & 4.90 & 4.89 & 4.96 & 4.71 & 4.06 & 3.16 & 4.74 \\
c4ai-command-a-03-2025 & 3.72 & 4.98 & 4.44 & 3.73& 4.39 & 3.16 & 4.88 & 4.62 & 4.93 & 4.48 & 4.08 & 2.94 & 4.47 \\
gpt-oss-120b & 3.58 & 5.00 & 4.46 & 3.62& 4.36 & 3.10 & 4.89 & 4.62 & 4.87 & 4.69 & 4.00 & 2.78 & 4.35 \\
gemma-3-27b-it & 3.62 & 5.00 & 4.35 & 3.54& 3.59 & 3.25 & 4.79 & 4.50 & 4.86 & 4.16 & 4.03 & 2.60 & 4.72 \\
aya-expanse-32b & 3.51 & 5.00 & 4.50 & 3.50& 4.09 & 3.64 & 4.75 & 4.38 & 4.89 & 4.76 & 3.91 & 2.72 & 4.60 \\ \bottomrule
\end{tabular}%
}
\caption{Human Evaluation Semantic Adequacy (1-5) across different countries}
\label{tab:semantic_adequacy_results}
\end{table*}
\begin{table*}[h]
\centering
\resizebox{\textwidth}{!}{%
\begin{tabular}{lccccccccccccc}
\toprule
\textbf{Model} & \textbf{EG} & \textbf{JO} & \textbf{LB} & \textbf{LY} & \textbf{MA} & \textbf{MR} & \textbf{OM} & \textbf{PS} & \textbf{SA} & \textbf{SD} & \textbf{SY} & \textbf{TN} & \textbf{YE} \\ \midrule
gemini-3-flash-preview\_medium & 3.56 & 4.87 & 4.57 & 4.05& 4.35 & 3.01 & 4.55 & 4.56 & 4.63 & 4.77 & 3.92 & 3.24 & 4.49 \\
gemini-3-flash-preview\_minimal & 3.62 & 4.83 & 4.54 & 3.95& 4.19 & 3.04 & 4.41 & 4.56 & 4.55 & 4.61 & 3.82 & 3.20 & 4.28 \\
c4ai-command-a-03-2025 & 3.31 & 3.95 & 4.14 & 3.30& 3.78 & 2.27 & 4.22 & 4.12 & 4.36 & 4.23 & 3.80 & 2.85 & 3.66 \\
gpt-oss-120b & 3.22 & 3.41 & 3.85 & 2.83& 3.30 & 2.13 & 4.36 & 3.73 & 4.15 & 3.28 & 3.61 & 2.58 & 3.67 \\
gemma-3-27b-it & 3.35 & 3.90 & 3.85 & 3.07& 3.04 & 2.08 & 4.17 & 3.86 & 4.26 & 3.69 & 3.66 & 2.30 & 3.86 \\
aya-expanse-32b & 3.02 & 3.15 & 3.37 & 2.78& 2.76 & 2.04 & 3.71 & 3.48 & 3.26 & 2.36 & 3.51 & 2.22 & 2.25 \\ \bottomrule
\end{tabular}%
}
\caption{Human Evaluation Dialectness \& Fluency (1-5) across different countries}
\label{tab:dialectness_fluency_results}
\end{table*}

\begin{figure}[t]
    \centering
    \includegraphics[width=0.49\textwidth]{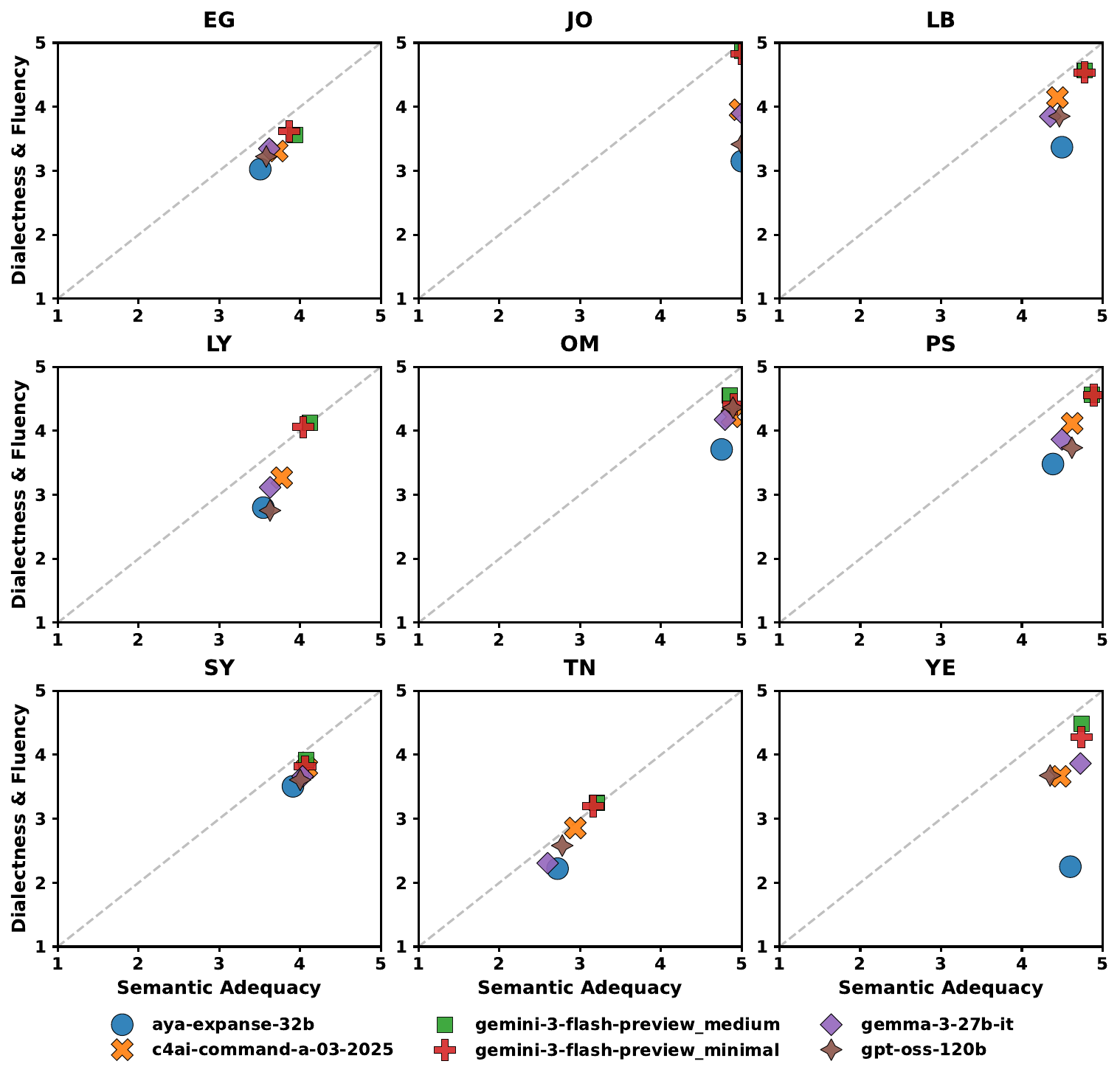}
    \caption{Human evaluation of Semantic Adequacy vs. Dialectness across the remaining dialects other than the ones presented in the main text. Points below the diagonal ($y=x$) indicate that models consistently achieve higher semantic fidelity than dialectal authenticity.}
    \label{fig:human_app_results}
\end{figure}

\end{document}